\documentclass[sigconf]{acmart}

\AtBeginDocument{%
  }

\copyrightyear{2025}
\acmYear{2025}
\setcopyright{acmlicensed}\acmConference[MM '25]{Proceedings of the 33rd
ACM International Conference on Multimedia}{October 27--31, 2025}{Dublin,
Ireland}
\acmBooktitle{Proceedings of the 33rd ACM International Conference on
Multimedia (MM '25), October 27--31, 2025, Dublin, Ireland}
\acmDOI{10.1145/3746027.3754821}
\acmISBN{979-8-4007-2035-2/2025/10}

\usepackage{amsfonts}
\usepackage{multirow}
\usepackage[most]{tcolorbox}
\usepackage{amsmath}

\usepackage{mdframed} 
\usepackage{booktabs}
\usepackage{subcaption}
\usepackage{colortbl}
\usepackage{algorithm}
\usepackage{algorithmic}

\begin{document}

\title{Court of LLMs: Evidence-Augmented Generation via Multi-LLM Collaboration for Text-Attributed Graph Anomaly Detection}

\settopmatter{printacmref=true, authorsperrow=4}
\author{Yiming Xu}
\authornote{Shaanxi Provincial Key Laboratory of Big Data Knowledge Engineering, Xi’an Jiaotong University, Xi’an, 710049, China}
\affiliation{
  \institution{School of Computer Science and Technology, Xi'an Jiaotong University}
  \city{Xi'an}
  \country{China}}
\email{xym0924@stu.xjtu.edu.cn}

\author{Jiarun Chen}
\authornotemark[1]
\affiliation{
  \institution{School of Computer Science and Technology, Xi'an Jiaotong University}
  \city{Xi'an}
  \country{China}}
\email{cjr519@stu.xjtu.edu.cn}

\author{Zhen Peng}
\authornotemark[1]
\authornote{Corresponding authors.}
\affiliation{
  \institution{School of Computer Science and Technology, Xi'an Jiaotong University}
  \city{Xi'an}
  \country{China}}
\email{zhenpeng27@outlook.com}

\author{Zihan Chen}
\affiliation{
  \institution{Department of Electrical and Computer Engineering, University of Virginia}
  \city{Charlottesville}
  \country{USA}}
\email{brf3rx@virginia.edu}

\author{Qika Lin}
\affiliation{
  \institution{Saw Swee Hock School of Public Health, National University of Singapore}
  \country{Singapore}}
\email{qikalin@foxmail.com}

\author{Lan Ma}
\affiliation{
  \institution{China Telecom Corporation Ltd. Shaanxi Branch}
  \city{Xi'an}
  \country{China}}
\email{malan@189.cn}

\author{Bin Shi}
\authornotemark[1]
\authornotemark[2]
\affiliation{
  \institution{School of Computer Science and Technology, Xi'an Jiaotong University}
  \city{Xi'an}
  \country{China}}
\email{shibin@xjtu.edu.cn}

\author{Bo Dong}
\authornotemark[1]
\affiliation{
  \institution{School of Distance Education, Xi'an Jiaotong University}
  \city{Xi'an}
  \country{China}}
\email{dong.bo@xjtu.edu.cn}

\renewcommand{\shortauthors}{Xu et al.}

\begin{abstract}
The natural combination of intricate topological structures and rich textual information in text-attributed graphs (TAGs) opens up a novel perspective for graph anomaly detection (GAD). However, existing GAD methods primarily focus on designing complex optimization objectives within the graph domain, overlooking the complementary value of the textual modality, whose features are often encoded by shallow embedding techniques, such as bag-of-words or skip-gram, so that semantic context related to anomalies may be missed. To unleash the enormous potential of textual modality, large language models (LLMs) have emerged as promising alternatives due to their strong semantic understanding and reasoning capabilities. Nevertheless, their application to TAG anomaly detection remains nascent, and they struggle to encode high-order structural information inherent in graphs due to input length constraints. For high-quality anomaly detection in TAGs, we propose \textbf{CoLL}, a novel framework that combines LLMs and graph neural networks (GNNs) to leverage their complementary strengths. CoLL employs multi-LLM collaboration for evidence-augmented generation to capture anomaly-relevant contexts while delivering human-readable rationales for detected anomalies. Moreover, CoLL integrates a GNN equipped with a gating mechanism to adaptively fuse textual features with evidence while preserving high-order topological information. Extensive experiments demonstrate the superiority of CoLL, achieving an average improvement of 13.37\% in AP. This study opens a new avenue for incorporating LLMs in advancing GAD. \footnote{The code and data are available at: https://github.com/yimingxu24/CoLL.} 
\end{abstract}



\received{20 February 2007}
\received[revised]{12 March 2009}
\received[accepted]{5 June 2009}

\maketitle

\section{Introduction}
\begin{figure}
  \centering
  \setlength{\belowcaptionskip}{-0.5cm}
  \captionsetup{aboveskip=5pt}
  \includegraphics[width=1\linewidth]{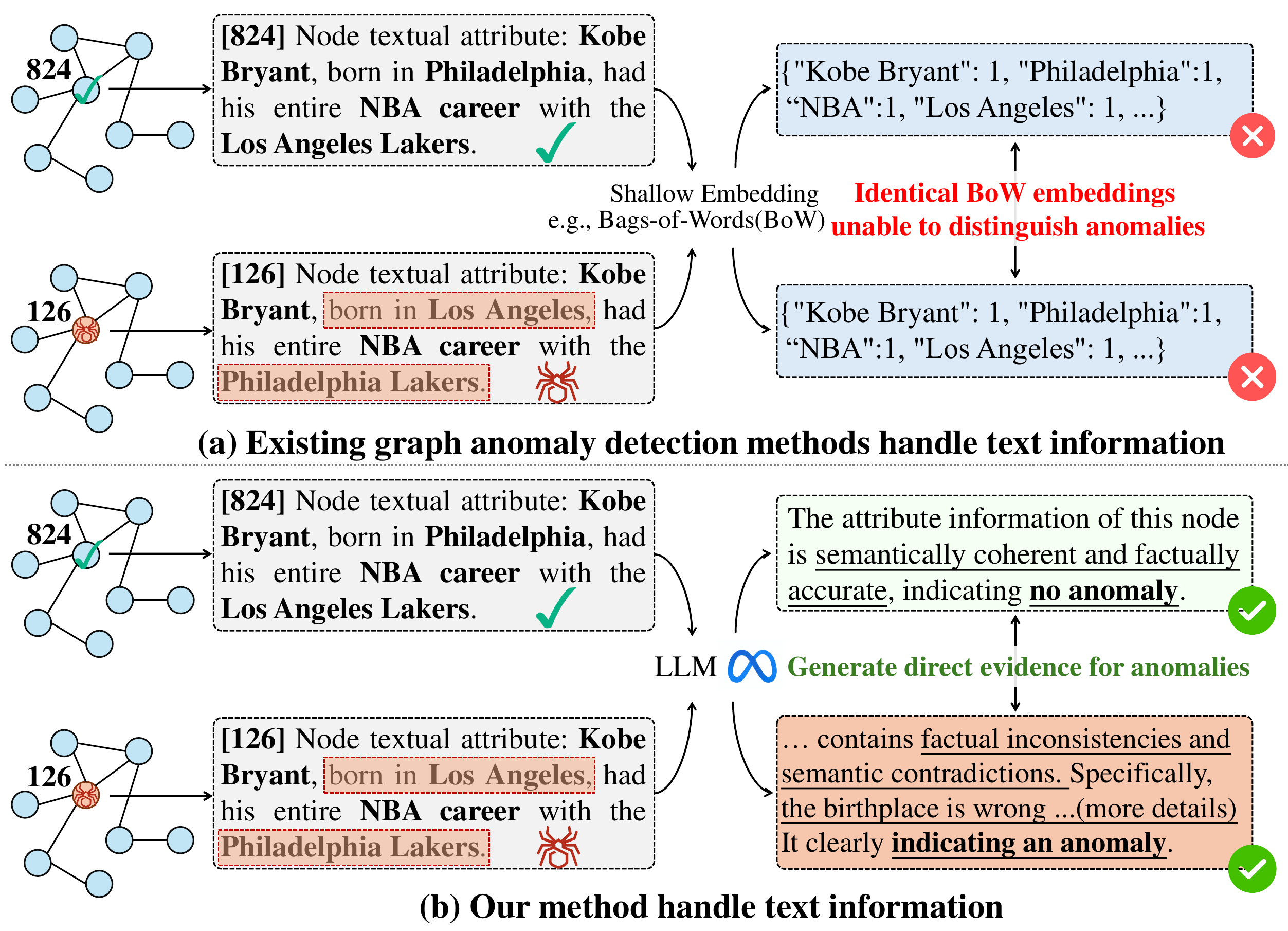}
  \caption{Illustration of our basic idea.}
  \label{fig:example}
\end{figure}

The superiority of graph data in capturing complicated interactions between entities has made it widely used in various high-impact domains, such as biomedicine~\cite{ye2021fenet}, cybersecurity~\cite {pazho2023survey}, and tax analysis~\cite{zheng2023survey}.
As reliability and stability become crucial in these graph-centric applications~\cite{cheng2021causal,zhang2022efraudcom,shi2023edge,xu2025ted}, there has been a growing focus on the graph anomaly detection (GAD)~\cite{ma2021comprehensive} task, which aims to identify suspicious entities that deviate significantly from normal.
In the real world, graph data often goes beyond structural interactions and involves rich textual information. For instance, e-commerce networks include rich raw texts in product names, descriptions, tags, and reviews~\cite{zhu2021knowledge}, while social media platforms encompass user profiles, posts, and comments~\cite{luo2017social}. Such graphs are commonly referred to as text-attributed graphs (TAGs)~\cite{zhang2024text,yan2023comprehensive}. 
Due to the natural involvement of both textual and structural modalities, TAGs offer a more nuanced perspective on anomalies, introducing the emerging challenge of text-attributed graph anomaly detection (TAGAD). \looseness=-1

While TAGs provide valuable textual signals beyond graph topology, existing efforts have predominantly focused on addressing the challenge of limited anomaly labels in the graph domain~\cite{liu2021anomaly}. To this end, researchers have developed elaborate self-supervised tasks for training graph neural networks (GNNs) to detect anomalies~\cite{duan2023graph,xu2025revisiting}. For example, SAMCL~\cite{hu2023samcl} employs six distinct loss functions to improve detection performance. However, the text information attached to nodes in TAGs rarely receives special attention. Most existing methods encode raw texts as shallow embeddings by multi-hot vector, bag-of-words (BoW)~\cite{harris1954distributional} or skip-gram~\cite{mikolov2013distributed} models for use. Despite notable progress, a key issue persists: existing textual feature extraction processes focus on learning general semantic patterns, which are inherently misaligned with the objectives of anomaly detection. As shown in Figure~\ref{fig:example}(a), node 126 contains obvious factual errors (wrong birthplace and team name, etc.), but shares the identical BoW encoding as the normal node 824. Since general-purpose encoding cannot expose anomaly-indicative signals hidden in the text semantics, the model struggles to identify subtle contextual irregularities. While fine-tuning general-purpose pretrained language models with anomaly labels might help bridge this gap, the sparsity of anomaly labels and the unbounded, diverse nature of anomalies in graphs~\cite{ma2021comprehensive} make it infeasible to train such large models in a reliable and scalable way.
Thus, effectively exposing anomaly cues from textual information without relying on labels remains an open challenge.

Fortunately, large language models (LLMs) successfully encapsulate extensive knowledge and have demonstrated remarkable capabilities across various knowledge-intensive tasks~\cite{zhao2023survey,yu2023kola}. Their strong semantic understanding and reasoning abilities open up new opportunities for generating direct, interpretable evidence or conclusion to support anomaly detection, particularly in capturing anomaly-specific contextual knowledge and subtle semantic inconsistencies. Analogous to presenting direct evidence in a courtroom, this approach minimizes distractions from irrelevant background information, allowing for correct and fair judgments~\cite{morgan2008impact,keane2022modern}, as illustrated in Figure~\ref{fig:example}(b).
In addition, insights from previous GAD research also emphasize the significance of capturing high-order information within graphs for precise anomaly detection~\cite{jin2021anemone,ma2021comprehensive}. 
However, directly feeding a node along with its multi-hop neighborhood textual information into an LLM to generate reliable evidence or accurate anomaly predictions faces significant challenges. As the neighborhood expands exponentially, the length and complexity of the associated textual context grow rapidly. Constrained by the limited input context length of LLMs~\cite{wang2024beyond} and their tendency to lose middle information when accessing long-context inputs~\cite{liu2024lost,firooz2024lost}, LLMs struggle to encode global high-order structural information~\cite{zhao2022learning,chen2023label}, which leads to suboptimal anomaly detection performance. In contrast, GNNs excel at preserving high-order topological structures with high fidelity~\cite{waikhom2023survey}, providing a complementary solution to address this limitation. Thus, it seems more sensible to combine the semantic understanding capabilities of LLMs with the topological modeling strengths of GNNs to address the inherent challenges of TAGAD.

To achieve high-quality detection capabilities tailored to TAGs, we propose \textbf{CoLL}, a novel framework for TAGAD.
As anomalies typically manifest as either contextual or structural~\cite{ma2021comprehensive}, CoLL leverages multi-LLM collaboration, where each LLM is assigned a specialized role, to generate evidence from complementary perspectives. In addition, CoLL integrates a gating mechanism and a GNN module to capture global high-order structural information embedded in the graph topology.
Specifically, we devise two LLM-driven prosecutors to generate evidence from contextual and structural perspectives.
The contextual prosecutor examines the factual accuracy of the textual attributes of a node, while the structural prosecutor assesses the consistency between the textual attributes of the target node and its adjacent neighbors. All candidate evidence is then consolidated and reviewed by a larger LLM acting as a judge, which delivers the final verdict. This collaborative prompting approach mitigates the drawback of individual LLM autocratic outputs and shows competitive performance compared to existing GAD methods. Interestingly, LLM-generated evidence is presented as human-readable rationales, further enhancing interpretability beyond traditional black-box GAD methods.
To mitigate the degradation caused by the lack of high-order structural information in LLMs, CoLL introduces a GNN equipped with a gating mechanism. The gating mechanism fuses the original textual attributes with LLM-generated verdicts into anomaly-aware representations, and feeds them into the GNN to preserve structural dependencies in the graph.
Finally, the model adopts a local inconsistency mining objective (node-subgraph contrast) to perform self-supervised training and assess the abnormality of nodes. Our main contributions are summarized as follows: \looseness=-1

$\bullet$ \textbf{\textit{Innovative Perspective}}: To the best of our knowledge, this work pioneers the incorporation of LLM responses to address the challenges of capturing anomaly-specific contextual and semantic knowledge in TAGAD, thereby opening new avenues for advancing anomaly detection.

$\bullet$ \textbf{\textit{Novel Algorithm}}: We propose CoLL, a novel framework inspired by courtroom dynamics, where LLMs act as prosecutors and judges, explicitly generating anomaly-related evidence and verdicts to bolster anomaly detection. Moreover, we incorporate gating mechanisms and GNNs to extract anomaly-relevant semantics and address the loss of high-order structural information in LLMs. 

$\bullet$ \textbf{\textit{Experimental Evaluation}}: CoLL outperforms 11 baselines on 4 text-attributed graph datasets, improving AUC by 2.39\% and AP by 13.37\% on average. Extensive ablation studies validate the contributions of each component. CoLL strikes a strong balance between accuracy and efficiency, while case studies (Appendix) highlight its superior interpretability over existing GAD methods.

\section{Related Work}
\subsection{Anomaly Detection in Attributed Graph}
Due to the labor-intensive nature of graph labeling, GAD methods typically adopt unsupervised paradigms~\cite{ma2021comprehensive}. Early approaches primarily rely on non-deep learning methods, such as matrix factorization~\cite{li2017radar,bandyopadhyay2019outlier} and clustering techniques~\cite{xu2007scan,muller2013ranking}. 
More recently, the rapid advancement of GNNs propels GAD into the deep learning era~\cite{ma2021comprehensive,xu2024learning}, offering enhanced performance in detecting graph anomalies.
DOMINANT~\cite{ding2019deep}, ADA-GAD~\cite{he2024ada}, and GAD-NR~\cite{roy2024gad} leverage graph autoencoders to measure node anomaly scores by leveraging the reconstruction errors. CoLA~\cite{liu2021anomaly} and ANEMONE~\cite{jin2021anemone} introduce contrastive learning techniques for GAD. Building on this, Sub-CR~\cite{zhang2022reconstruction}, GRADATE~\cite{duan2023graph} and SAMCL~\cite{hu2023samcl} incorporate subgraph-level contrast for more accurate node anomaly score estimation. AEGIS~\cite{ding2021inductive} uses autoencoders and generative adversarial learning to identify anomalies. \cite{liu2024arc,lin2024unigad} attempt to develop a general framework. Despite recent progress, existing research predominantly focuses on designing complex self-supervised tasks for attributed graphs. However, the crucial challenges of mitigating irrelevant background introduced during text feature extraction and effectively leveraging the rich textual information in TAGs to enhance detection capabilities remain largely unexplored. \looseness=-1

\subsection{Graph Learning with LLMs}
With the rise of LLMs~\cite{achiam2023gpt,dubey2024llama,yang2024qwen2}, their powerful capabilities are transforming how we interact with graphs. Recent efforts to apply LLMs to TAGs demonstrate promising potential~\cite{zhang2023graph,jin2024large,mao2024advancing}. 
LLMs can serve as predictors in graph learning~\cite{huang2024can}. InstructGLM~\cite{ye2023language} uses natural language to describe the geometric structure of the graph, and then instruction finetunes an LLM to perform graph tasks. GraphText~\cite{zhao2023graphtext} derives a graph-syntax tree for each graph, utilizing an LLM to process the graph text sequences generated from traversing the tree. GraphGPT~\cite{tang2024graphgpt} integrates LLMs with structural knowledge through graph instruction tuning. 
However, recent studies~\cite{huang2024can} indicate that current LLMs interpret input prompts merely as linearized text rather than genuinely understanding the underlying graph structures.
In addition, LLMs can also act as enhancers, leveraging their strengths to efficiently boost performance of smaller models. GIANT~\cite{chien2021node} leverages XR-Transformers~\cite{zhang2021fast} and can output better feature vectors than bag-of-words and vanilla BERT~\cite{devlin2018bert} for node classification. OFA~\cite{liu2023one} describes graphs in natural language and uses LLM to unify inputs to build graph foundation models. Despite significant progress, existing methods overlook the potential of collaborative LLMs, and remain focused on supervised settings and node classification, with limited exploration of unsupervised GAD. Our work takes a first step toward filling this gap. \looseness=-1

\begin{figure*}
  \centering
  \setlength{\belowcaptionskip}{-0.3cm}
  \includegraphics[width=1\linewidth]{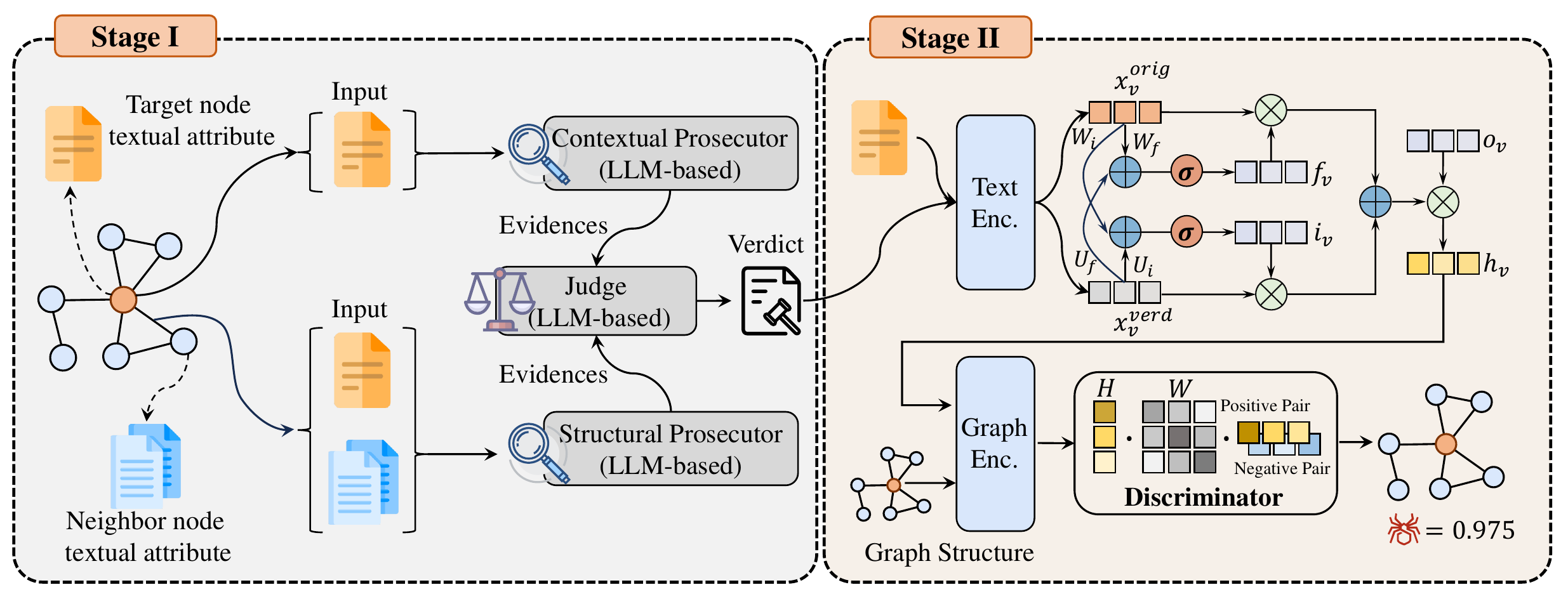}
  \caption{The overview of CoLL. Stage I: Evidence-Augmented Generation. Stage II: High-order Information Completion.}
  \label{fig:overview}
\end{figure*}

\section{Methodology}
In this section, we first formally define the TAGAD task and then introduces CoLL, an unsupervised TAGAD framework that integrates LLM-based evidence-augmented generation and GNN-based high-order information completion. 
An overview of the CoLL workflow is presented in Figure~\ref{fig:overview}.

\subsection{Problem Formulation}
\textbf{Text-Attributed Graph}.
A text-attributed graph (TAG) is defined as  $\mathcal{G}= \left ( \mathcal{V},\mathcal{E}, \mathcal{T},\mathbf{A} \right )$, where $\mathcal{V}$ represents the set of nodes, $\mathcal{E}$ denotes the set of edges. $\mathcal{T} = \{t_v \mid v \in \mathcal{V}\}$ is the set of textual attributes associated with each node, where $t_{v} \in \mathcal{D}^{L_v}$, with $\mathcal{D}$ representing the dictionary of words or tokens, and $L_v$ denoting the sequence length of node $v$. The adjacency matrix $\mathbf{A}\in \left\{ 0,1\right\}^{n\times n}$ indicates graph connectivity where $\mathbf{A}\left [ i,j \right ]=1$ represents an edge between nodes $i$ and $j$ in $\mathcal{E}$. TAG contains structural information from $(\mathcal{V},\mathcal{E})$ and textual information from $\mathcal{T}$ to perform various graph-related downstream tasks.

\textbf{Unsupervised Text-Attributed Graph Anomaly Detection}.
Given a TAG $\mathcal{G}$, the objective is to learn an anomaly score function $f: \mathcal{V} \rightarrow \mathbb{R}$ to detect nodes that significantly deviate from the majority of nodes, without access to labeled anomalies during the training process. The function $f(v_i)$ assigns an anomaly score to each node $v_i \in \mathcal{V}$, with higher scores indicating a greater likelihood of the node being anomalous.

\subsection{Evidence-Augmented Generation}
\paragraph{Evidence Generation by Prosecutor}
Existing GAD methods~\cite{ding2019deep,liu2021anomaly} typically yield textual representations dominated by general semantics, overlooking anomaly-relevant signals.
Inspired by the human judicial system~\cite{morgan2008impact,keane2022modern}, we are motivated to explore using LLMs to simulate courtroom trials, generating direct evidence indicating whether a node is anomalous or normal. This approach aims to compensate for the limitations of existing text encoders in capturing anomaly-specific contextual knowledge and semantic understanding. However, the unique characteristics and inherent complexity of GAD make generating evidence with LLMs a non-trivial task.

Since anomalies are typically categorized as contextual and structural anomalies~\cite{ma2021comprehensive}, we utilize human-readable natural language prompts to effectively guide and instruct LLMs to focus on anomaly detection from these two distinct perspectives. To this end, we design a contextual prosecutor and a structural prosecutor tailored to each anomaly type. Subsequent empirical experiments demonstrate that using separate LLMs dedicated to each perspective yields better results compared to employing a single LLM to handle both perspectives simultaneously.
Specifically, the goal of the contextual prosecutor is to evaluate whether the textual attributes associated with each node $i$ exhibit anomalies. To achieve this, we construct a prompt that incorporates the dataset context, the contextual textual features of the node, and detailed instructions to guide the LLM in assessing potential anomalies. This prompt is then provided to the contextual prosecutor, with detailed prompt formulations in Appendix~\ref{sec:appendix_prompt}. The general structure of the prompt is as follows:

\begin{tcolorbox}[boxsep=0mm,left=2.5mm,right=2.5mm,breakable]
\textbf{Instruction:} [after providing a brief introduction to the data sample, ask the prosecutor to evaluate whether the provided text contains any content anomalies, such as irrelevant or inconsistent information that deviates from its main topic or theme]\\
\textbf{Content:} [paper abstract or book description]\\
\textbf{Title:} [paper or book title if available]\\
\textbf{Instruction:} [provide evidence and conclude with either "normal" or "abnormal"]\\
\textbf{Answer:}
\label{box:instruction}
\end{tcolorbox}

For the structural prosecutor, given the constraint on input context length for LLMs, we perform a sampling process by selecting the first-order neighbors  $j \in \mathcal{N}(i)$ of the target node $i$. The contextual features of both the target node and its sampled neighbors are then fed into the structural prosecutor. The general structure of this input is as follows:

\begin{tcolorbox}[boxsep=0mm,left=2.5mm,right=2.5mm]
\textbf{Instruction:} [provide a text from one central node, along with the sampling texts of its neighbors sampled five times, ask the prosecutor to evaluate whether the central node has meaningful relationships with up to five sampled neighbors]\\
\textbf{Central node:} $<$\textit{node content}$>$: $<$\textit{node title}$>$\\
\textbf{Neighbor nodes:} $<$\textit{neighbor content 1}$>$: $<$\textit{neighbor title 1}$>$; $<$\textit{neighbor content 2}$>$: $<$\textit{neighbor title 2}$>$; $<$\textit{neighbor content 3}$>$: $<$\textit{neighbor title 3}$>$; ...\\
\textbf{Instruction:} [provide evidence and conclude with either "related" or "unrelated"]\\
\textbf{Answer:}
\end{tcolorbox}

The evidence (output) generated by both prosecutors is in a human-readable text format. Compared to the human-incomprehensible feature vectors produced by traditional GAD methods, this format offers better interpretability (further details can be found in the case study). The general format of the evidence is as follows:

\begin{tcolorbox}[boxsep=0mm,left=2.5mm,right=2.5mm]
(\textbf{Evidence}) [prosecutor-generated evidence for the prediction]\\
(\textbf{Prediction}) [a single-word prediction]
\end{tcolorbox}

\paragraph{Verdict Generation by Judge}
By combining the efforts of the contextual prosecutor and the structural prosecutor, we can gather evidence from two different perspectives. However, due to the gap between LLM generation and understanding~\cite{west2023generative}, there may be low quality or inconsistencies in the results~\cite{fu2023generate}. Previous studies have shown that LLMs have the preliminary ability to judge and evaluate their own answers~\cite{kadavath2022language,feng2024don}. 
To reduce the likelihood of misjudgments, we propose a novel multi-LLM collaborative framework that simulates a courtroom-like interaction.

Specifically, contextual and structural prosecutors independently generate their respective sets of evidence, i.e., each prosecutor produces multiple outputs tailored to its perspective. Then, we introduce a more powerful LLM as a judge, tasked with synthesizing the contextual information of the node (including textual and neighbor attributes) alongside all evidence from both prosecutors to deliver a final verdict. In this way, we aim to leverage collaboration and supervision among multiple LLMs to improve overall decision consistency and reliability, thereby enhancing performance in GAD tasks. The general prompt template input to the judge is as follows:

\begin{tcolorbox}[boxsep=0mm,left=2.5mm,right=2.5mm]
\textbf{Instruction:} [provide all prior prosecutors' conclusions regarding the central node. Ask the judge to review and evaluate these prior judgments and make a final decision on whether the central node is anomalous]\\
\textbf{Contextual evidence:} $<$\textit{node content}$>$; $<$\textit{contextual evidence 1}$>$; $<$\textit{contextual evidence 2}$>$; $<$\textit{contextual evidence 3}$>$ ...\\
\textbf{Structural evidence:} $<$\textit{neighbor content 1}$>$: $<$\textit{structural evidence 1}$>$; $<$\textit{neighbor content 2}$>$: $<$\textit{structural evidence 2}$>$; $<$\textit{neighbor content 3}$>$: $<$\textit{structural evidence 3}$>$ ...\\
\textbf{Instruction:} [ask the model to evaluate the conclusions of the prosecutors, provide supporting evidence and conclude with either "normal" or "abnormal"]\\
\textbf{Answer:}
\end{tcolorbox}

The output format of the judge is similar to that of the prosecutors, maintaining a human-readable and interpretable structure. Before arriving at the final judgment, the judge carefully evaluates the outputs provided by the prosecutors, offering evidence to support its final decision. This evidence serves as a concise explanation of the rationale behind the judge's judgment.

\begin{tcolorbox}[boxsep=0mm,left=2.5mm,right=2.5mm]
(\textbf{Evidence}) [an evaluation of certain prosecutors' conclusions as evidence to support the judge's own decision]\\
(\textbf{Judgment}) [a single-word judgment]
\end{tcolorbox}

\subsection{High-order Information Completion}
Through the collaboration of multiple LLMs, we obtain a final verdict in natural language form, including specific explanations regarding whether a node is anomalous and the rationale behind it. The LLM-based courtroom achieves competitive performance compared to GAD-specific baselines, while additionally offering more interpretable verdicts as a byproduct. However, anomalies often manifest at different scales within a graph~\cite{jin2021anemone}, and the limited input context of LLMs hinders their ability to capture high-order structural information, leading to performance bottlenecks. To address this limitation, we propose a GNN equipped with a gating mechanism, aiming to enable more robust anomaly detection.
\paragraph{Anomaly-Aware Feature Fusion by Gating Mechanism}
First, we use a frozen pre-trained text encoder to transform the original text $\mathcal{T}_{\textrm{orig}}$ and the final verdict $\mathcal{T}_{\textrm{verd}}$ produced by LLMs into node features that are suitable for downstream GNNs, as illustrated below:
\begin{equation}
\label{eq:textenc}
    \begin{split}
        \mathbf{x}_v^\textrm{orig} = f_{\text{text}}(t_v^{\textrm{orig}}), \quad
        \mathbf{x}_v^\textrm{verd} = f_{\text{text}}(t_v^{\textrm{verd}}),
    \end{split}
\end{equation}
where $t_v^{\textrm{orig}}$ and $t_v^{\textrm{verd}}$ are the original text and generated verdict of node $v$. $f_{\text{text}}$ represents the text encoder. We obtain the node text features $\mathbf{x}_v^\textrm{orig}$ and verdict features $\mathbf{x}_v^\textrm{verd}$ through $f_{\text{text}}$.

While LLM-generated verdicts offer valuable task-relevant signals, the original textual information also encodes essential semantic cues. To fully exploit both sources, we incorporate feature fusion to enhance the anomaly-aware representation for anomaly detection.
Inspired by the LSTM architecture~\cite{hochreiter1997long}, we devise three gating mechanisms: a forget gate, an input gate, and an output gate to facilitate feature fusion. The purpose of the forget gate is to determine which parts of the original text features are irrelevant to the anomaly and should be selectively forgotten. Meanwhile, the input gate controls which valuable information from the newly generated verdict features should be retained. The output gate determines which features are ultimately output. Each gate simultaneously considers both the original textual features and the verdict features, enabling a more precise and effective fusion of information for anomaly detection. The formulas are as follows: 
\begin{equation}
\label{eq:gate1}
    \begin{aligned}
        \mathbf{f}_v &=  \sigma\left ( \mathbf{W}_f \cdot \mathbf{x}_v^\textrm{orig} + \mathbf{U}_f \cdot \mathbf{x}_v^\textrm{verd} + \mathbf{b}_f \right ),\\
        \mathbf{i}_v &= \sigma\left (\mathbf{W}_i \cdot \mathbf{x}_v^\textrm{orig} +  \mathbf{U}_i \cdot \mathbf{x}_v^\textrm{verd} + \mathbf{b}_i \right ),\\
        \mathbf{o}_v &= \sigma\left (\mathbf{W}_o \cdot \mathbf{x}_v^\textrm{orig} +  \mathbf{U}_o \cdot \mathbf{x}_v^\textrm{verd} + \mathbf{b}_o \right ),
    \end{aligned}
\end{equation}
where $\mathbf{f}_v$, $\mathbf{i}_v$, and $\mathbf{o}_v$ represent the activation vectors of the forget gate, input gate, and output gate for node $v$, respectively. $\sigma$ denotes the Sigmoid activation function and layer normalization. $\mathbf{W}_*$ and $\mathbf{U}_*$ are the trainable weight matrices for each gate, while $\mathbf{b}_*$ are the corresponding bias terms.

Finally, the original textual features and verdict features are integrated through the forget and input gates, while the output gate determines which information from the fused representation is propagated into the GNN. These gating mechanisms regulate the information flow at each step, enabling the model to combine evidence effectively, selectively retain anomaly-relevant rationales, and filter out irrelevant noise signals. The updated node representation, which serves as the input to the GNN, is computed as follows: \looseness=-1
\begin{equation}
\label{eq:gate2}
    \begin{aligned}
        \mathbf{\hat{x}}_v &= \mathbf{f}_v \cdot \mathbf{x}_v^\textrm{orig} + \mathbf{i}_v \cdot \mathbf{x}_v^\textrm{verd},\\
        \mathbf{h}_v &= \mathrm{LN} \left ( \mathbf{o}_v \cdot \mathbf{\hat{x}}_v \right ),
    \end{aligned}
\end{equation}
where $\mathbf{h}_v$ is the fusion feature node by gating mechanism. $ \mathrm{LN} \left ( \cdot \right )$ is layer normalization. 

\paragraph{Graph Contrastive Network}
After feature fusion, we obtain anomaly-relevant numerical features that are compatible with GNNs. To capture anomalies at different scales, we design node-subgraph contrastive pairs to train the model, enabling it to learn the neighborhood matching relationships of primarily normal nodes in the graph, adhering to the homophily assumption.
First, to exploit the structural modeling capacity of GNNs, we input both the node features and the structural information of the TAG into the GNNs. The node representations $\mathbf{Z}$ are obtained by the GNN module:
\begin{equation}
\label{eq:gnn}
\mathbf{Z} = f_{\text{gnn}}\left ( \mathbf{A},\mathbf{H} \right ),
\end{equation}
where $f_{\text{gnn}}$ represents the graph encoder. $\mathbf{A}$ is the adjacency matrix and $\mathbf{H}$ is the node feature after feature fusion. Then, we compute the subgraph representations using the \textrm{Readout} function, which has been widely used in previous work~\cite{hassani2020contrastive,xu2023cldg}:
\begin{equation}
\mathbf{e}_{v} = \text{Readout}(\tilde{\mathbf{Z}}_{v}) = \sum_{k=1}^{ \left| \mathcal{N}\left ( v \right ) \right|} \frac{\mathbf{Z}_k}{ \left| \mathcal{N}\left ( v \right ) \right|},
\end{equation}
where $\mathbf{e}_{v}$ represents the subgraph representation of node $v$, $\tilde{\mathbf{Z}}_{v}$ denotes the neighbor feature matrix of node $v$, and $\left| \mathcal{N}(v) \right|$ indicates the number of neighbors for node $v$.

Subsequently, we apply a discriminator to calculate the similarity score $s_{v}$ between the node-subgraph pairs. This is achieved through a bilinear scoring function, as follows:
\begin{equation}
\label{eq:disc}
s_{v}=\textrm{Bilinear}\left ( \mathbf{e}_v, \mathbf{h}_{v} \right )=\sigma \left ( \mathbf{e}_v\mathbf{W} \mathbf{h}_{v} \right ),
\end{equation}
where $\mathbf{W}$ is a trainable matrix, and $\sigma\left ( \cdot  \right ) $ is $\textrm{Sigmoid}$ function.

The objective of the discriminator is to accurately learn the neighborhood matching relationships within the graph, effectively distinguishing between the relationships of a node and its own neighbors (positive pairs) and those with other nodes' neighbors (negative pairs). To achieve this, we use the binary cross-entropy (BCE) loss~\cite{velivckovic2018deep} as the objective function:
\begin{equation}
\label{eq:loss}
\mathcal{L}=-\frac{1}{2 n} \sum_{i=1}^n\left(\log \left(s_{v}^{+}\right)+\log \left(1-s_{v}^{-}\right)\right),
\end{equation}
where $\left(s_{v}^{+}\right)$ and $\left(s_{v}^{-}\right)$ are the positive and negative similarities of node $v$, respectively. The parameters of the gating mechanism, GNN, and discriminator are updated by minimizing $\mathcal{L}$.

\subsection{Anomaly Score Inference}
By minimizing the objective function above, the model is trained to learn the topological relationships of a large number of normal nodes. However, anomalous nodes, whether anomalous in terms of features or structure~\cite{ma2021comprehensive}, often exhibit local inconsistency~\cite{liu2021anomaly}, meaning they are dissimilar to both positive and negative pairs. Thus, for a given node $v$, we define its anomaly score by the similarity scores of both positive and negative pairs as follows:
\begin{equation}
\label{eq:inf}
f_{\textrm{score}}\left ( v \right )=\frac{\sum_{r=1}^{R}\left ( s_{v}^{-}-s_{v}^{+} \right )}{R},
\end{equation}
where $f_{\textrm{score}}\left ( v \right )$ represents the final anomaly score for node $v$, with a higher score indicating a greater likelihood of being anomalous. $R$ denotes the number of sampling rounds. A detailed summary of the CoLL workflow is provided in Algorithm~\ref{alg:coll} in the Appendix.

\begin{table}\huge
\centering
\renewcommand{\arraystretch}{1.2}
\setlength{\tabcolsep}{4pt}
\caption{Statistics of the Datasets.}
\label{tab:dataset_stats}
\resizebox{\linewidth}{!}{
\begin{tabular}{lcccc}
\toprule
\textbf{Dataset} & \textbf{Nodes} & \textbf{Edges} & \textbf{Avg. Doc Length} & \textbf{Anomalies} \\
\midrule
Cora       & 2,708   & 10,984    & 135.45    & 108   \\ 
Pubmed     & 19,717  & 90,368    & 256.08    & 788   \\
History    & 41,551  & 369,252   & 228.36    & 1,662 \\
ogbn-Arxiv & 169,343 & 1,210,112 & 179.70    & 6,774 \\
\bottomrule
\end{tabular}
}
\end{table}

\begin{table*}[t]
\centering
\setlength{\tabcolsep}{4pt}
\caption{Experimental results for TAGAD on four datasets (OOM: Out of Memory). We report mean AUC and AP. \textbf{Bold} indicates the best result, and the runner-up is \underline{underlined}.}
\resizebox{\textwidth}{!}{
\begin{tabular}{ccccccccc}
\toprule
\multirow{2}{*}{\textbf{Method}} & \multicolumn{2}{c}{\textbf{Cora}} & \multicolumn{2}{c}{\textbf{Pubmed}} & \multicolumn{2}{c}{\textbf{History}} & \multicolumn{2}{c}{\textbf{ogbn-Arxiv}} \\
\cmidrule(lr){2-3} \cmidrule(lr){4-5} \cmidrule(lr){6-7} \cmidrule(lr){8-9}
& \textbf{AUC} & \textbf{AP} & \textbf{AUC} & \textbf{AP} & \textbf{AUC} & \textbf{AP} & \textbf{AUC} & \textbf{AP} \\
\midrule
LOF 
& 55.11$_{\pm 0.00}$ & 4.45$_{\pm 0.00}$ 
& 64.10$_{\pm 0.00}$ & 8.39$_{\pm 0.00}$ 
& 55.71$_{\pm 0.00}$ & 5.22$_{\pm 0.00}$ 
& 71.85$_{\pm 0.00}$ & 19.53$_{\pm 0.00}$\\
SCAN 
& 56.77$_{\pm 0.73}$ & 4.70$_{\pm 0.01}$
& 60.16$_{\pm 0.52}$ & 10.41$_{\pm 0.64}$
& 57.27$_{\pm 0.89}$ & 5.00$_{\pm 0.70}$
& 58.98$_{\pm 0.81}$ & 5.48$_{\pm 0.73}$ \\

Radar 
& 51.17$_{\pm 0.33}$ & 3.83$_{\pm 0.00}$
& 51.88$_{\pm 0.35}$ & 3.92$_{\pm 0.24}$
& 48.48$_{\pm 1.05}$ & 3.60$_{\pm 0.76}$
& OOM & OOM \\
\midrule

AEGIS 
& 51.01$_{\pm 2.42}$ & 4.33$_{\pm 0.30}$
& 49.08$_{\pm 1.10}$ & 3.66$_{\pm 0.50}$
& 52.28$_{\pm 1.97}$ & 4.23$_{\pm 0.75}$
& 54.03$_{\pm 1.44}$ & 4.41$_{\pm 0.60}$ \\

MLPAE 
& 51.23$_{\pm 0.87}$ & 4.41$_{\pm 0.07}$
& 50.56$_{\pm 0.26}$ & 3.93$_{\pm 0.39}$
& 47.54$_{\pm 0.16}$ & 3.67$_{\pm 0.53}$
& 51.17$_{\pm 0.50}$ & 3.91$_{\pm 0.23}$ \\

DOMINANT 
& 63.71$_{\pm 1.04}$ & 8.09$_{\pm 0.17}$
& 67.32$_{\pm 0.61}$ & 7.54$_{\pm 0.81}$
& 63.46$_{\pm 0.23}$ & 6.17$_{\pm 0.15}$
& OOM & OOM \\

GAD-NR 
& 68.70$_{\pm 1.51}$ & 9.65$_{\pm 0.47}$
& 66.25$_{\pm 0.44}$ & 6.34$_{\pm 0.04}$
& 65.86$_{\pm 0.13}$ & 5.89$_{\pm 0.22}$
& 65.23$_{\pm 0.51}$ & 6.86$_{\pm 0.15}$ \\
\midrule

CoLA 
& 77.69$_{\pm 0.99}$ & 14.95$_{\pm 1.48}$
& $\underline{78.83_{\pm 0.55}}$ & 18.93$_{\pm 0.62}$
& 78.72$_{\pm 0.09}$ & 26.48$_{\pm 0.55}$
& 81.03$_{\pm 0.15}$ & $\underline{27.73_{\pm 0.40}}$ \\

ANEMONE 
& $\underline{80.03_{\pm 0.72}}$ & 18.60$_{\pm 1.47}$
& 78.81$_{\pm 0.02}$ & $\underline{19.43_{\pm 0.54}}$
& $\underline{79.14_{\pm 0.02}}$ & $\underline{27.20_{\pm 0.66}}$
& 80.97$_{\pm 0.03}$ & 27.49$_{\pm 0.75}$ \\

SL-GAD 
& 76.51$_{\pm 0.96}$ & $\underline{21.18_{\pm 0.43}}$
& 75.19$_{\pm 0.35}$ & 16.71$_{\pm 0.05}$
& 74.08$_{\pm 0.28}$ & 20.22$_{\pm 0.41}$
& $\underline{81.16_{\pm 0.21}}$ & 27.64$_{\pm 0.41}$ \\

GRADATE 
& 74.91$_{\pm 0.81}$ & 15.05$_{\pm 1.02}$
& 67.49$_{\pm 0.09}$ & 12.13$_{\pm 0.18}$
& 73.25$_{\pm 0.10}$ & 14.50$_{\pm 0.55}$
& OOM & OOM \\
\midrule
\rowcolor[HTML]{E9E9E9} 
\textbf{CoLL} 
& $\mathbf{80.27_{\pm 0.89}}$ & $\mathbf{27.35_{\pm 1.20}}$
& $\mathbf{81.93_{\pm 0.58}}$ & $\mathbf{37.04_{\pm 1.54}}$
& $\mathbf{79.34_{\pm 0.22}}$ & $\mathbf{39.66_{\pm 2.36}}$
& $\mathbf{87.17_{\pm 0.63}}$ & $\mathbf{44.95_{\pm 2.07}}$ \\
\bottomrule
\end{tabular}}
\label{tab:ad}
\end{table*}

\subsection{Discussion}
Existing GAD methods often rely on shallow text encodings and focus on designing complex optimization objectives to boost detection performance~\cite{zhang2022reconstruction,duan2023graph}.
Our key contribution lies in highlighting an overlooked challenge: without anomaly-aware textual encoding, as illustrated by the toy example in Figure~\ref{fig:example}, existing methods may be prone to the "garbage in, garbage out" problem. This suggests that advances in GAD may be constrained not by optimization strategies, but by the quality and relevance of input representations.

In this work, we focus on leveraging LLMs to directly extract anomaly-relevant cues from the textual modality. While recent works have explored applying LMs to graph learning~\cite{he2023harnessing,liu2023one,huang2023can}, they primarily target semi-supervised node classification tasks, emphasizing node features while overlooking the potential of multi-LLM collaboration and the modeling of graph topology. In contrast, we propose an unsupervised framework featuring a courtroom-inspired multi-LLM collaboration scheme, where two prosecutors provide complementary evidence from contextual and structural perspectives, and a judge synthesizes their reasoning to reach a final verdict. Additionally, we introduce an adaptive gating mechanism that selectively preserves anomaly-indicative rationales from both raw textual attributes and LLM-generated verdicts, while a GNN module captures high-order structural information. CoLL demonstrates superior advantages over existing GAD methods from multiple perspectives, including significantly improved detection performance (Section~\ref{sec:main_res}), scalability (Section~\ref{sec:time}), and enhanced interpretability (Appendix~\ref{sec:appendix_case}). We believe this paradigm shift will open a new direction for advancing anomaly detection in graphs.

\section{Experiments}
\subsection{Experimental Setup}

\paragraph{Datasets.}
We conduct comprehensive experiments on four TAG datasets, each comprising both graph structures and textual attributes associated with nodes. Cora, Pubmed, and ogbn-Arxiv are citation networks~\cite{sen2008collective,hu2020open}, while History is an e-commerce network~\cite{yan2023comprehensive}. Each node is labeled with one of two labels: normal or abnormal. The statistics of these datasets are summarized in Table~\ref{tab:dataset_stats}. Detailed descriptions of all datasets are provided in Appendix~\ref{sec:appendix_datasets}.

\vspace{-0.2cm}
\paragraph{Baselines.} 
We compare our method against six categories of SOTA unsupervised anomaly detection baselines, including the density-based model: LOF~\cite{breunig2000lof}. Structural clustering-based model: SCAN~\cite{xu2007scan}. Matrix factorization-based model: Radar~\cite{li2017radar}. Generative adversarial learning-based model: AEGIS~\cite{ding2021inductive}. Reconstruction-based models: MLPAE~\cite{sakurada2014anomaly}, DOMINANT~\cite{ding2019deep} and GAD-NR~\cite{roy2024gad}. Contrastive learning-based models: CoLA~\cite{liu2021anomaly}, AENMONE~\cite{jin2021anemone}, SL-GAD~\cite{zheng2021generative}, and GRADATE~\cite{duan2023graph}.

\vspace{-0.2cm}
\paragraph{Implementation.} 
We select Llama 3.1 8B as the contextual and structural prosecutors and Llama 3.1 70B as the judge~\cite{dubey2024llama}. The text encoder in Eq.\eqref{eq:textenc} is implemented using BGE\cite{bge_embedding}. For fairness, the same BGE is used to encode raw node text into numerical features for baselines that cannot directly process textual data. The Adam optimizer~\cite{kingma2014adam} is utilized, and the learning rates, epochs, batch size, GNN layer and L2 regularization for the gating and GNN components are set as follows: Cora (3e-3, 25, 256, 2, 1e-4), Pubmed (5e-4, 100, 512, 2, 1e-4), History (5e-3, 25, 512, 2, 0.0), and ogbn-Arxiv (5e-3, 100, 256, 2, 1e-3). $R$ is set to 256 and the hidden dimension is 64. More details are provided in Appendix~\ref{sec:appendix_param}.

\vspace{-0.2cm}
\subsection{Main Results} \label{sec:main_res}
To comprehensively evaluate the performance of our proposed method, we conduct experiments on four datasets for TAGAD under an unsupervised setting, comparing against 11 baseline models. We employ ROC-AUC and average precision (AP) as evaluation metrics, as they are widely used and well-suited for anomaly detection tasks.

As shown in Table~\ref{tab:ad}, CoLL consistently surpasses all prior methods on the TAGAD task. Shallow methods, such as LOF, SCAN, and Radar, struggle to capture the complex features and structural anomalies inherent in TAGs. Reconstruction-based methods demonstrate partial improvements, yet their reliance on full graph reconstruction often renders them impractical for large-scale scenarios.  Contrastive learning methods achieve state-of-the-art results by designing complex loss functions in all baselines. However, all existing approaches overlook a critical point: general-purpose text encoders encode generic knowledge and introduce noise unrelated to anomalies. 

To overcome this limitation, we integrate evidence-augmented generation from LLMs, focusing on anomaly-specific cues to enhance detection. Appendix~\ref{sec:appendix_case} provides four representative case studies showing the effectiveness of LLM collaboration in generating high-quality anomaly-specific evidence while showing stronger interpretability. Leveraging high-order information capture of GNN, CoLL achieves improvements of 2.39\% in AUC and 13.37\% in AP compared to the runner-up method. This work opens a promising pathway for leveraging LLMs in TAGAD.

\begin{figure*}[t]
    \centering
    \setlength{\belowcaptionskip}{-0.2cm}
    \subfloat[Cora]{\label{fig:roc1}\includegraphics[width=0.25\linewidth]{
    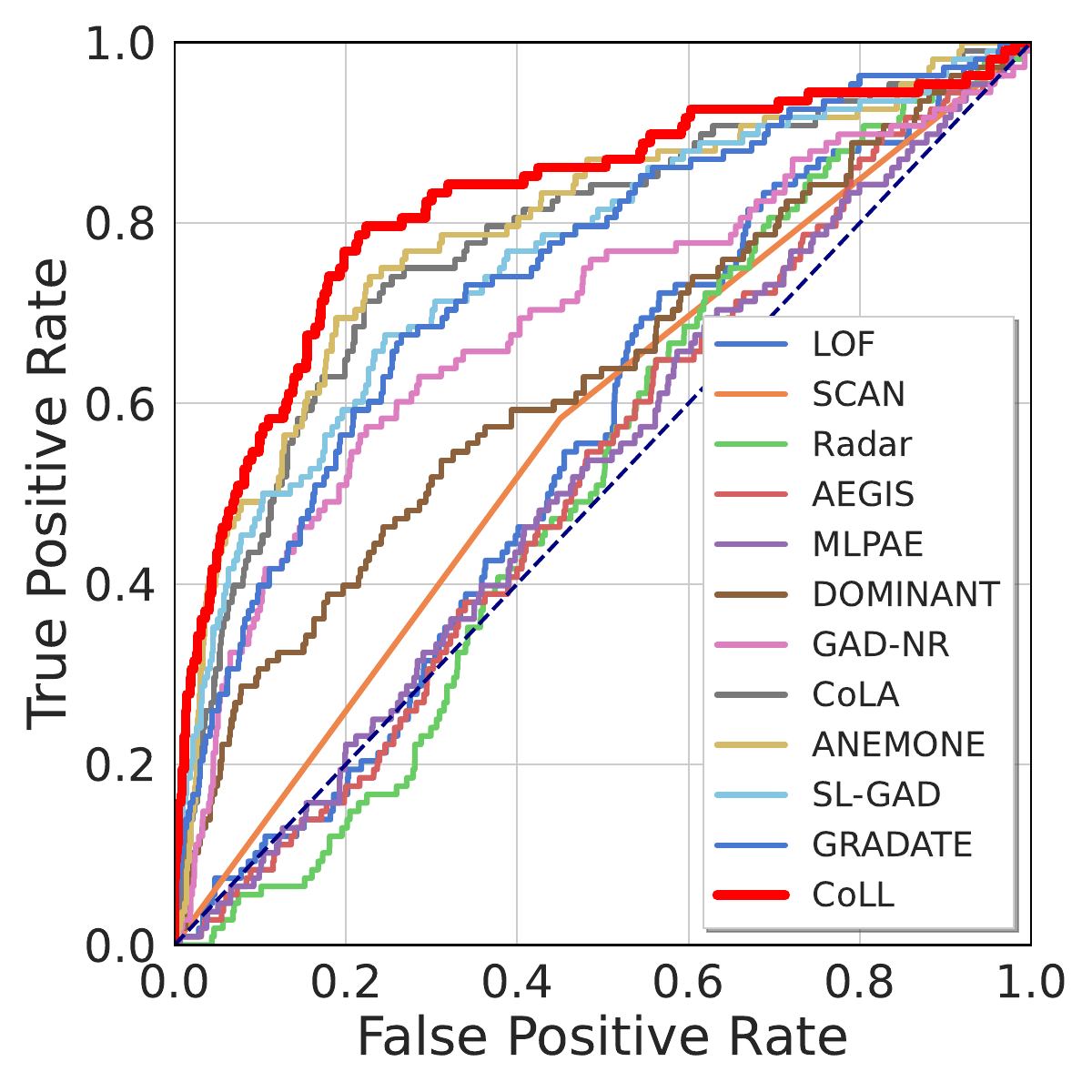}}
    \subfloat[Pubmed]{\label{fig:roc2}\includegraphics[width=0.25\linewidth]{
    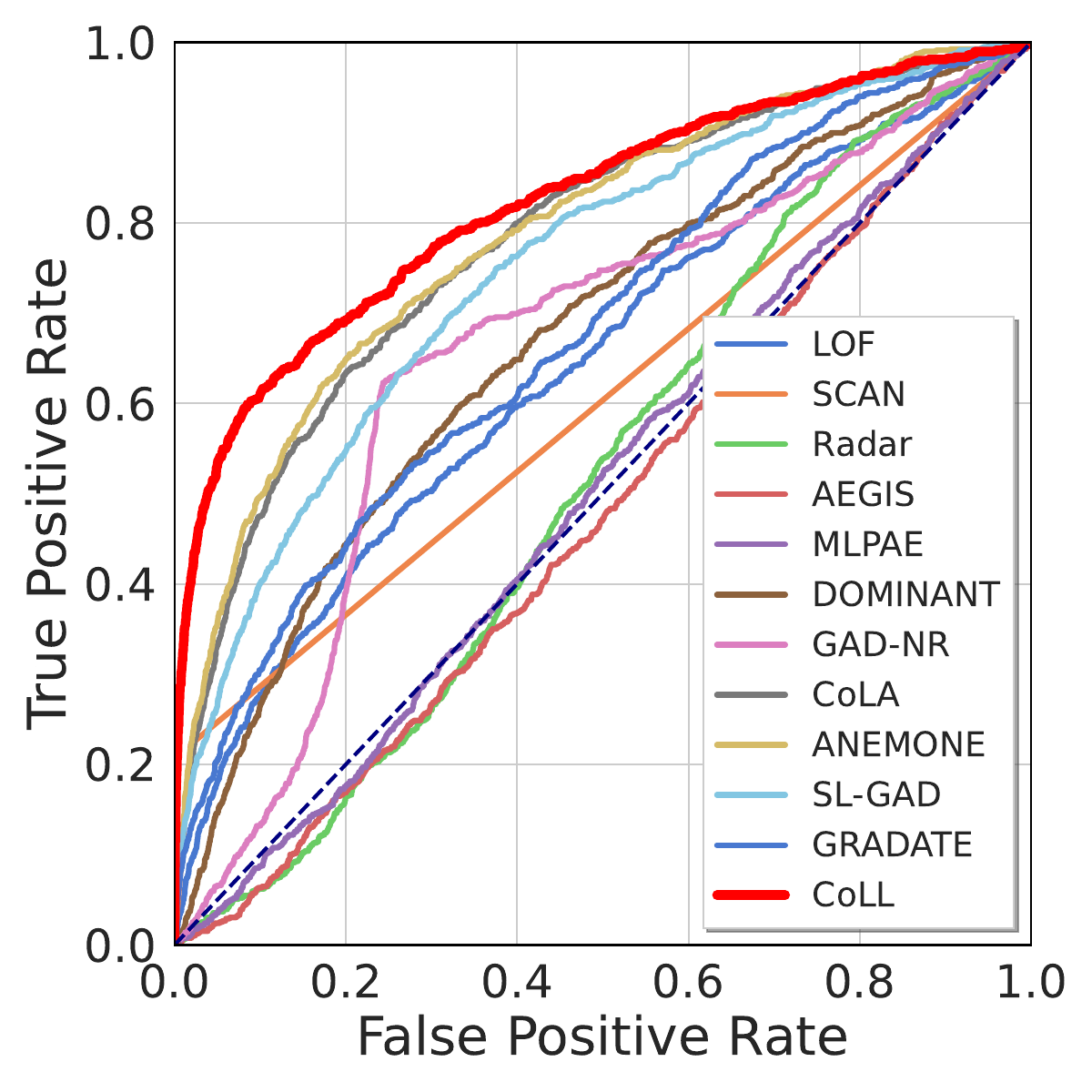}}
    \subfloat[History]{\label{fig:roc3}\includegraphics[width=0.25\linewidth]{
    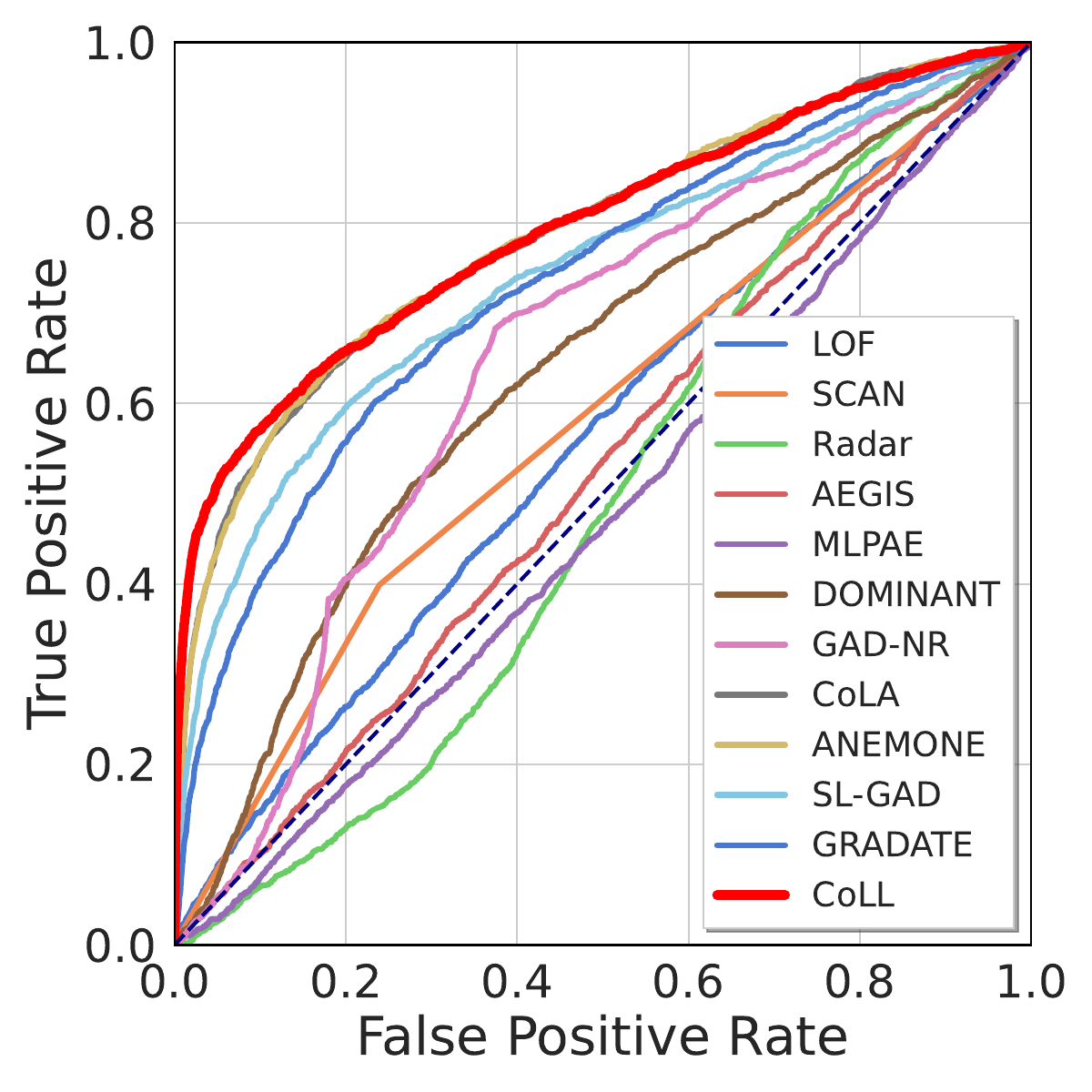}}
    \subfloat[ogbn-Arxiv]{\label{fig:roc4}\includegraphics[width=0.25\linewidth]{
    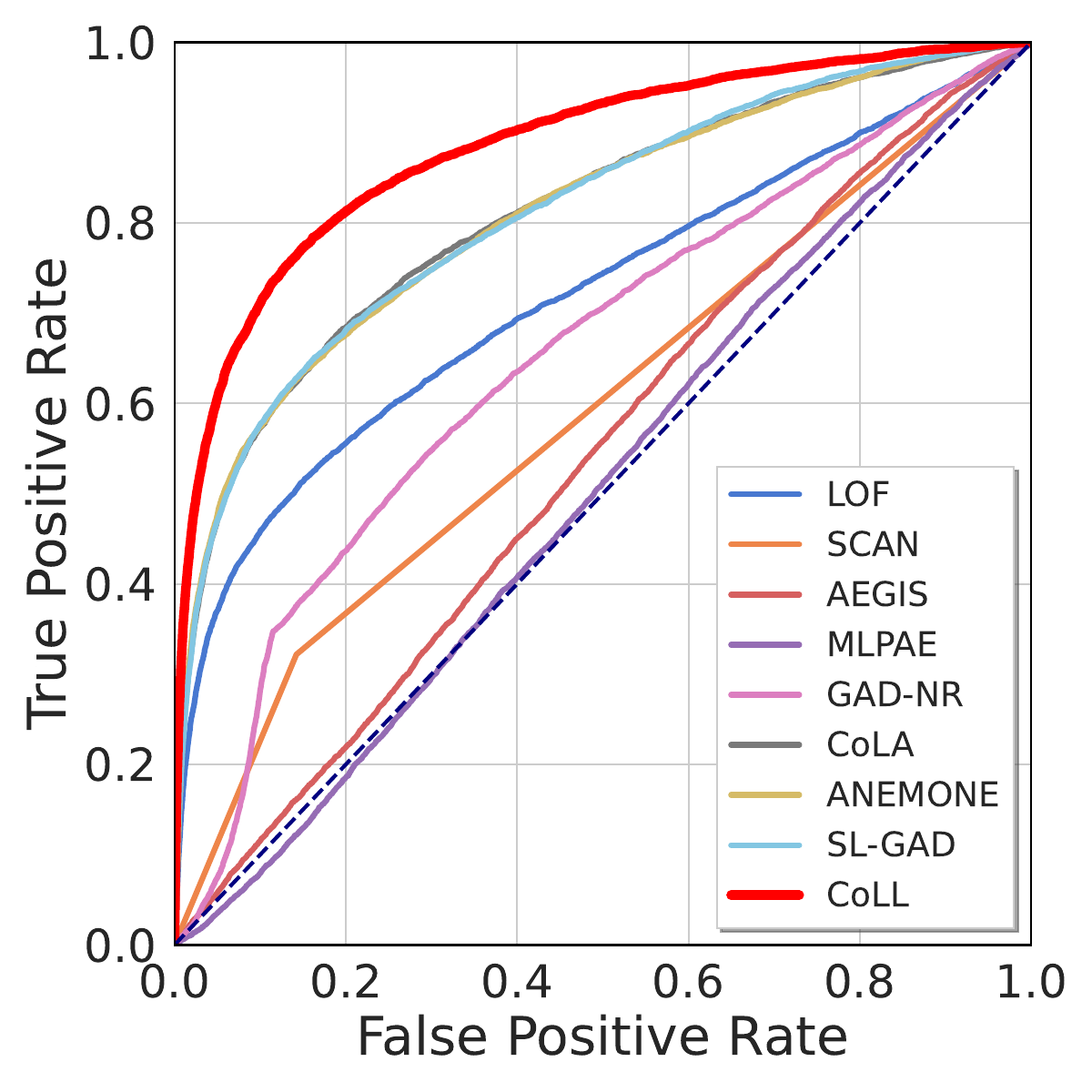}}
    \caption{ROC curves across four datasets. A larger area under the curve indicates better performance. The black dashed lines represent the performance of random guessing.}
    \label{fig:roc}    
\end{figure*}

\begin{figure*}[t]
\centering
\setlength{\belowcaptionskip}{-0.24cm}
\subfloat[Stage I]{\label{fig:abl1}\includegraphics[width=0.45\linewidth]{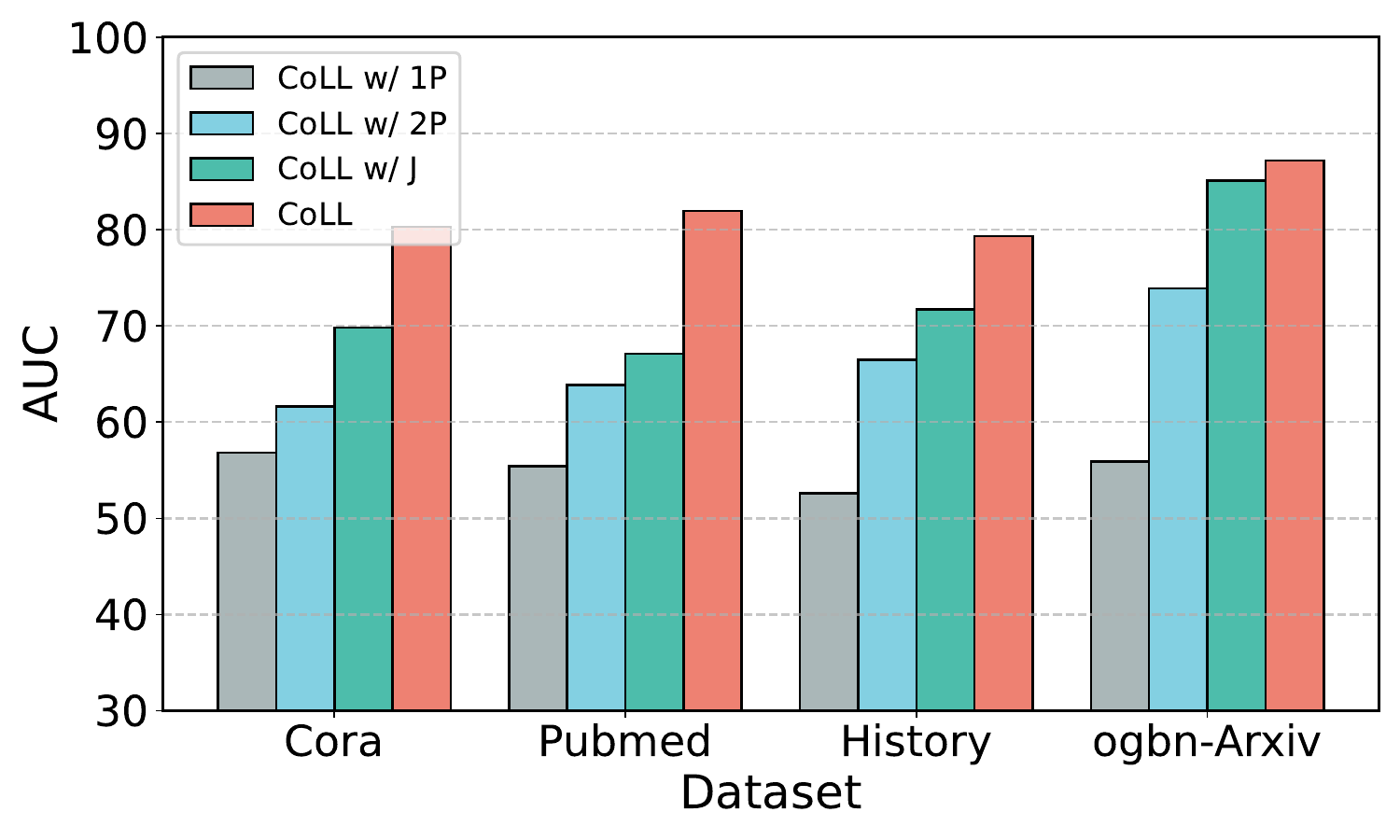}}
\subfloat[Stage II]{\label{fig:abl2}\includegraphics[width=0.45\linewidth]{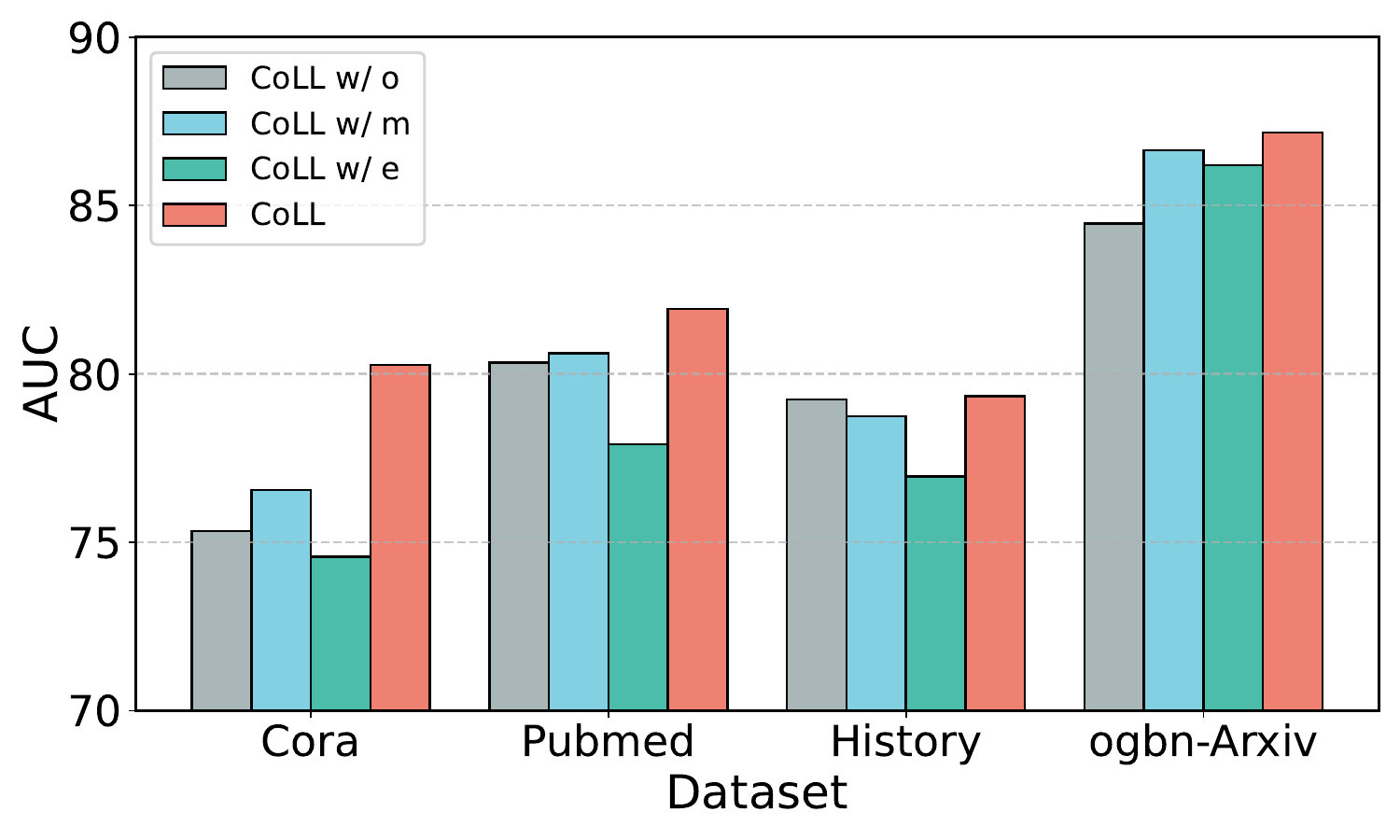}}
\caption{The ablation study result w.r.t. AUC.}
\label{fig:abl}
\end{figure*}

\subsection{Ablation Study}  \label{sec:abl}
We conduct comprehensive ablation studies to evaluate the effectiveness of the proposed components. We first focus on the effectiveness of the LLM court in Stage I. Figure~\ref{fig:abl1} presents the results of three CoLL variants: CoLL \textrm{w/ 1P}, where a single prosecutor evaluates both contextual and structural anomalies; CoLL \textrm{w/ 2P}, where two prosecutors independently assess contextual and structural anomalies; and CoLL \textrm{w/ J}, which incorporates an additional judge to synthesize and refine evidence. None of these variants include the gating mechanism or GNN components.

We begin by providing both structural and attribute information to CoLL \textrm{w/ 1P}, aiming to leverage LLM capabilities and integrate information. However, the results fall short of expectations, highlighting the limitations of LLMs in handling long-range dependencies introduced by high-order information. Recent studies have also demonstrated that large language models tend to lose middle information when processing long-context inputs~\cite{liu2024lost,firooz2024lost}.
Therefore, we attempt to refine the anomaly identification process by employing two prosecutors, each focusing on a distinct perspective: one capturing contextual information and the other analyzing structural patterns. 
The consistent improvements across all datasets confirm the benefits of disentangling anomalies using specialized prosecutors (CoLL \textrm{w/ 2P}), outperforming the single-prosecutor design.
This finer-grained approach aims to enhance detection performance by disentangling and leveraging complementary information from both views. Furthermore, introducing a judge (CoLL \textrm{w/ J}) significantly enhances performance, highlighting the utility of integrating multiple perspectives to refine evidence and improve anomaly detection. The high-order information supplemented by the GNN further improves the performance. These findings validate the contribution of each component in achieving robust and effective results. 

Subsequently, we conduct a comprehensive ablation study specifically targeting the Stage II gating mechanism. Figure~\ref{fig:abl2} illustrates the results of four different methods for utilizing the raw node features and evidence (or verdict) features: CoLL \textrm{w/o}: Only the raw node features are fed into the GNN. CoLL \textrm{w/m}: The raw node features and the LLM-generated verdict features are directly combined using mean pooling before being input into the GNN. CoLL \textrm{w/e}: Only the LLM-generated verdict features are used as input to the GNN. None of the above three variants contains a gating mechanism. CoLL: Our full framework employs the gating mechanism to fuse both types of features before feeding them into the GNN.

The experimental results show that CoLL, equipped with the gating feature fusion mechanism, consistently achieves the best performance. This highlights that the gating mechanism, optimized for anomaly detection objectives, effectively preserves anomaly-relevant semantic information. CoLL \textrm{w/m}, which averages the two feature types, also outperforms CoLL \textrm{w/o} in most cases, confirming the benefit of evidence augmentation. The performance of CoLL \textrm{w/ e} aligns with the observations in Figure~\ref{fig:abl}, indicating that as the performance of CoLL \textrm{w/ J} improves, the quality of the evidence features also increases. 

Overall, extensive experiments validate the contribution of each component, including multi-LLM collaboration and the GNN equipped with the gating mechanism.

\subsection{Parameter Study}
\begin{figure}
\centering
\setlength{\belowcaptionskip}{-0.3cm}
\subfloat[sampling rounds $R$]
{\label{fig:param1}\includegraphics[width=0.5\linewidth]{
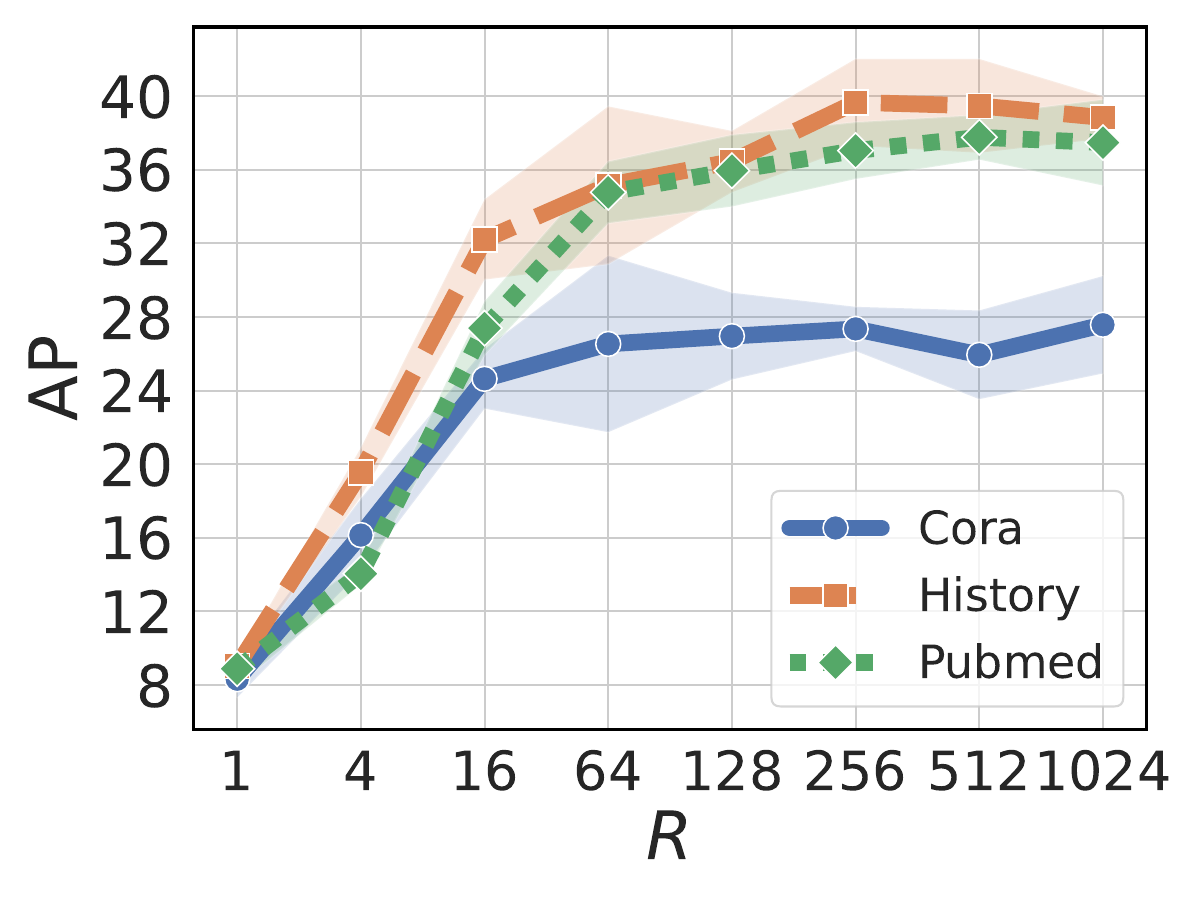}}
\subfloat[dimension $d$]{\label{fig:param2}\includegraphics[width=0.5\linewidth]{
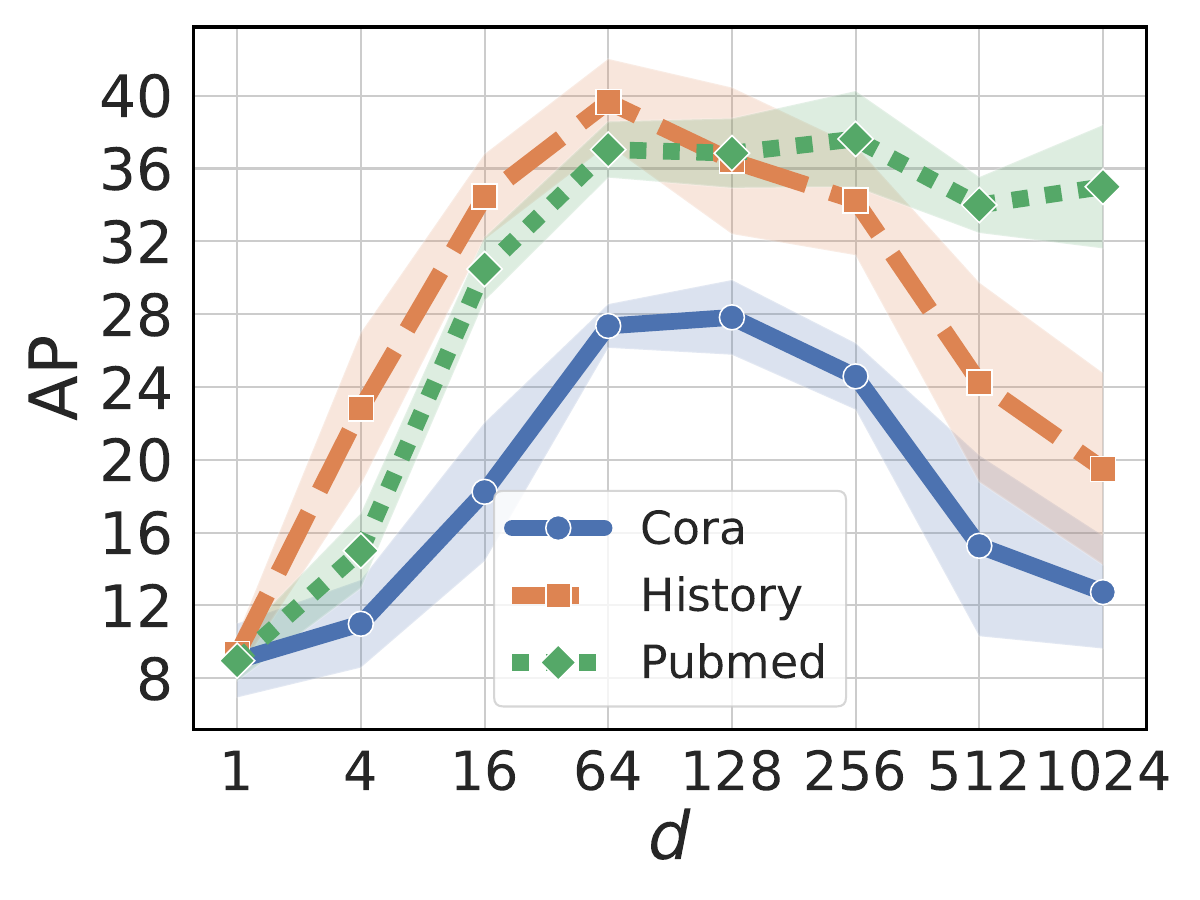}}
\caption{The parameter study result w.r.t. AP.}

\label{fig:param}
\end{figure}

\textbf{Effect of sampling rounds $R$}\quad We evaluate the impact of the sampling rounds $R$ in Eq.\eqref{eq:inf}. As shown in Figure~\ref{fig:param1}, the AP score improves as $R$ increases, since a small number of negative samples in limited batches may not provide sufficient discrimination. However, the performance stabilizes when $R$ exceeds 256, indicating diminishing returns with further increases. To balance performance and efficiency, we set $R=256$ as the default choice for all datasets.

\noindent \textbf{Effect of hidden dimension $d$}\quad We study the effect of hidden dimension $d$ on detection performance. When $d$ increases to 64, all datasets show improved performance. However, with further increase of $d$, the AP decreases significantly. The information related to the anomaly is specific and limited. Larger dimensions tend to preserve excessive semantic noise, leading to degraded performance. Thus, we recommend avoiding extreme feature dimensions. Empirically, we set $d = 64$ across all datasets for optimal performance. \looseness=-1

\subsection{Time Complexity and Cost Estimation} \label{sec:time}
In this subsection, we analyze the time complexity and cost estimation using the ogbn-Arxiv dataset (169,343 nodes, 1,210,112 edges), significantly larger than most graphs used in prior GAD studies~\cite{ding2019deep,liu2021anomaly, jin2021anemone, duan2023graph}. Our proposed CoLL framework comprises two main stages: Stage I (multi-LLM collaboration) and Stage II (gating and GNN inference). To quantify the computational overhead, we first focus on Stage I, as it represents the primary contributor to both time consumption and cost. Given the sensitivity of the data, we adopt a local server setup to mitigate the risk of data leakage at any stage of the process. However, significant differences in hardware configurations can impact the estimation of cost and inference time. To ensure generalizability, we compute the cost and time for Stage I based on the pricing and inference speed reported by Artificial Analysis~\footnote{https://artificialanalysis.ai}. On average, the input sequences for the contextual prosecutor and structure prosecutor (Llama 3.1 8B) consist of 317 and 410 tokens, respectively, while their outputs contain 47 and 124 tokens. The judge's input and output sequences consist of 2257 and 157 tokens, respectively. For Llama 3.1 8B API, the blended pricing is \$0.03 per million tokens, with an output speed of 2173 tokens per second. For Llama 3.1 70B API, the blended pricing is \$0.20 per million tokens, with the same output speed of 2173 tokens per second. The cost estimation for ogbn-Arxiv is as follows:
\begin{equation}
\resizebox{1\hsize}{!}{$
\textit{Cost} = \left( \frac{(317 + 410 + 47 + 124) \times 0.03}{10^6} + \frac{(2257 + 157) \times 0.2}{10^6} \right) \times 169,343 \approx 86.3 \ USD
$}
\end{equation}

Given that LLMs process input tokens in batch parallel computation with significantly higher efficiency compared to the autoregressive generation process, we focus solely on the time required for generating the output sequence, which is computed as follows:
\begin{equation}
\textit{Time} = \frac{47 + 124 + 157}{2173 \times 60} \times 169,343 \approx 426 \textit{min} \approx 7.1 \textit{h}
\end{equation}

Given the increasing complexity of modern LLMs, the cost and runtime of Stage I remain acceptable for million-scale graphs, especially considering its performance gains in anomaly detection. Moreover, while prosecutors and judges operate sequentially, inference across different nodes within each LLM can be parallelized. Assuming a parallelism factor of $p$, the runtime can be reduced to $7.1 \textit{h}/p$, making the approach highly scalable to larger graphs. Meanwhile, LLM-generated evidence and verdicts are stored for subsequent use, maximizing efficiency and reducing costs.

\begin{figure}
\centering
\setlength{\belowcaptionskip}{-0.3cm}
\includegraphics[width=1\linewidth]{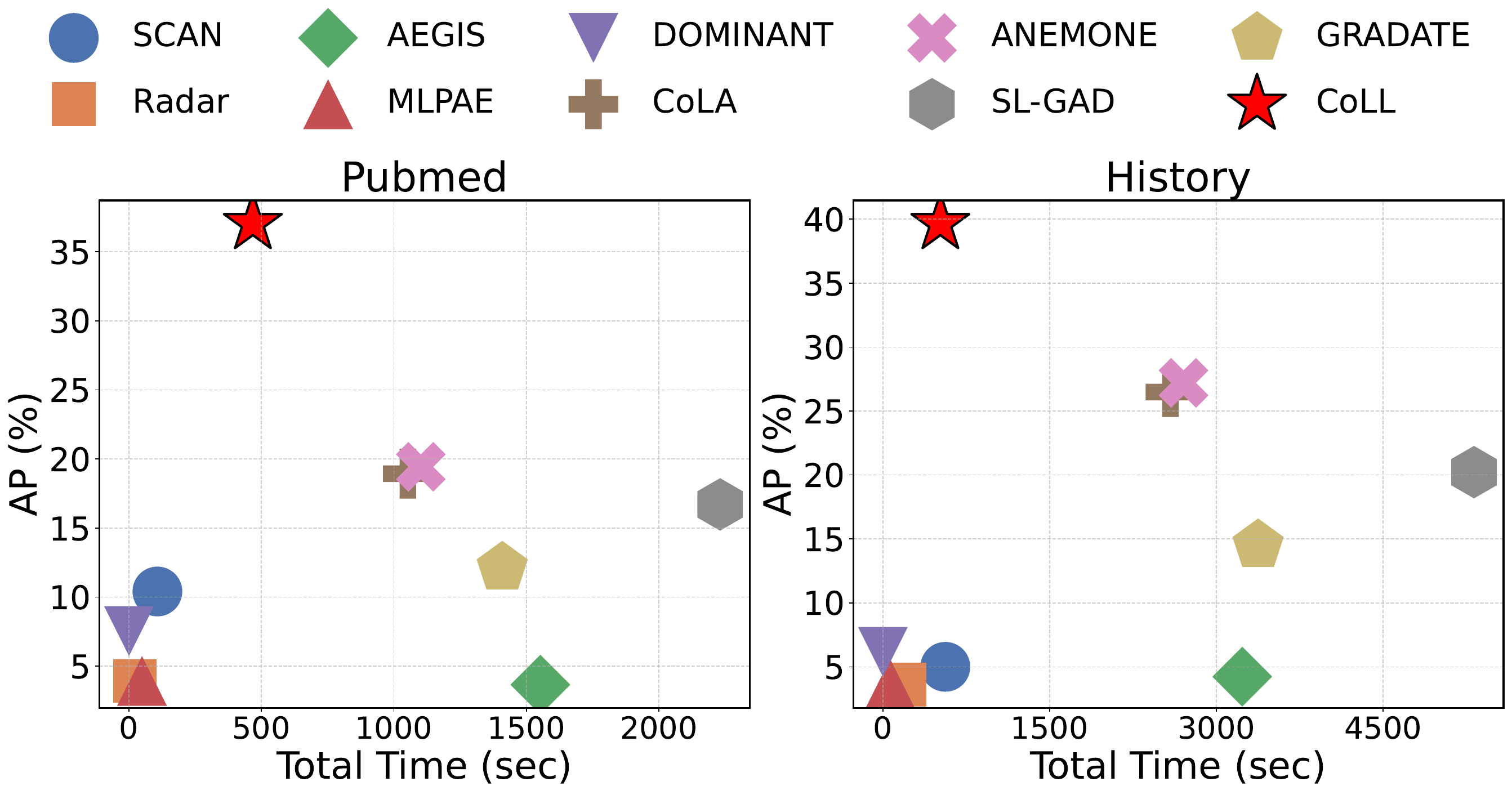}
\caption{The performance trade-off between anomaly detection capability and training time on Pubmed and History.}
\label{fig:time}
\end{figure}

The time complexity of Stage II primarily arises from gating and GNN computations. The gating mechanism has a complexity of $O\left(\left|\mathcal{V}\right|d^{2} \right)$, where $\mathcal{V}$ is the node set and $d$ is the feature dimension. The GNN operates with $O\left(\left|\mathcal{V}\right|d^{2}+\left| \mathcal{E}\right|d\right)$, where $\mathcal{E}$ is the edge set. Thus, the overall complexity of Stage II in CoLL is $O\left(\left|\mathcal{V}\right|d^{2}+\left| \mathcal{E}\right|d\right)$, which is comparable to existing graph contrastive learning-based anomaly detection methods.
As shown in Figure~\ref{fig:time}, CoLL achieves near-optimal runtime, trailing only behind some shallow methods. CoLL runs 9.37× faster than deep learning-based approaches while achieving a 13.37\% higher AP than the runner-up method. This demonstrates its strong balance between efficiency and detection performance, making it a highly practical solution for scalable anomaly detection.  The detailed running results can be found in Table~\ref{tab:runtime}.

\section{Conclusion}
This paper introduces CoLL, a novel framework for text-attributed graph anomaly detection (TAGAD), which seamlessly integrates LLMs for semantic reasoning and GNNs for high-order topological modeling. CoLL leverages multi-LLM collaboration to generate human-readable anomaly-related evidence from both contextual and structural perspectives. Through a gating mechanism and GNN integration, CoLL effectively captures anomaly-relevant semantics and high-order structural information. Experiments on four datasets validate CoLL’s superiority, surpassing all baselines and setting a new benchmark in TAGAD. This work highlights the potential of LLMs in advancing GAD by addressing existing limitations.


\begin{acks}
This research was partially supported by the Key Research and Development Project in Shaanxi Province No. 2023GXLH-024, the National Science Foundation of China No. 62302380, 62476215, 62037001, 62137002 and 62192781, and the China Postdoctoral Science Foundation No. 2023M742789.
\end{acks}

\bibliographystyle{ACM-Reference-Format}
\bibliography{sample-base}

\appendix
\clearpage

\begin{algorithm}[!t]
    \caption{CoLL}
    \label{alg:coll}
    \raggedright
    
    \textbf{Input}: Graph dataset $\mathcal{G}= \left ( \mathcal{V},\mathcal{E}, \mathcal{T},\mathbf{A} \right )$, initial gate parameter $\theta_{gate}$, GNN encoder parameter $\theta_{GNN}$, epoch $E$, learning rate $\eta_1,\eta_2$\\
    \textbf{Output}: Anomaly score function $f: \mathcal{V} \rightarrow \mathbb{R}$ 
    \begin{algorithmic}[1]
    \STATE \textcolor{gray}{================== Stage I =====================}    
    \STATE Compute contextual evidence $t_c^{\text{evi}}$ by processing $\mathcal{V}, \mathcal{T}$ along with predefined prompt $p_c$ using the contextual prosecutor
    \STATE Compute structural evidence $t_s^{\text{evi}}$ by processing $\mathcal{V}, \mathcal{E}, \mathcal{T}$ along with predefined prompt $p_s$ using the structural prosecutor 
    \STATE Compute the final verdict $t^{\text{verd}}$ by integrating $\mathcal{G}, t_c^{\text{evi}}, t_s^{\text{evi}}$ along with predefined prompt $p_j$ using the judge
    \STATE \textcolor{gray}{================== Stage II =====================}  
    \STATE $\mathbf{x}_v^\textrm{orig} = f_{\text{text}}(t_v^{\textrm{orig}})$
    \STATE $\mathbf{x}_v^\textrm{verd} = f_{\text{text}}(t_v^{\textrm{verd}})$
    \FOR{$t=1, \cdots, E$}
    \STATE $\mathbf{h}_v = f_{\textrm{gate}}(\mathbf{x}_v^\textrm{orig}, \mathbf{x}_v^\textrm{verd}; \theta_{gate})$, where $f_{\textrm{gate}}$ is defined in Eq.~\eqref{eq:gate1} and Eq.~\eqref{eq:gate2}
    \STATE $\mathbf{Z} = f_{\text{gnn}}\left ( \mathbf{A},\mathbf{H}; \theta_{GNN} \right )$ 
    \STATE $\mathbf{e}_{v} = \text{Readout}(\tilde{\mathbf{Z}}_{v})$
    \STATE Compute $s_{v}^{+}$ and $s_{v}^{-}$ by discriminator Eq.~\eqref{eq:disc}
    \STATE Calculate the loss $\mathcal{L}$ by Eq.~\eqref{eq:loss} using $s_{v}^{+}$ and $s_{v}^{-}$ 
    \STATE Update the GNN encoder parameters $\theta_{\text{GNN}} \leftarrow \theta_{\text{GNN}} - \eta_1 \nabla_{\theta_{\text{GNN}}} \mathcal{L}$
    \STATE Update the gate parameters \\ $\theta_{\text{gate}} \leftarrow \theta_{\text{gate}} - \eta_2 \nabla_{\theta_{\text{gate}}} \mathcal{L}$
    \ENDFOR
    \STATE Calculate the final anomaly score $f_{\textrm{score}}\left ( v; \theta_{gate},\theta_{GNN} \right )$ of node $v$ by Eq.~\eqref{eq:inf}
    \RETURN{$f_{\textrm{score}}\left (\mathcal{V} \right )$}
    \end{algorithmic}
\end{algorithm}

\section{Algorithm}
This section presents the algorithmic workflow of CoLL, as outlined in Algorithm~\ref{alg:coll}. CoLL is a framework for text-attributed graph anomaly detection (TAGAD), which consists of two stages: evidence-augmented generation (Stage I) and high-order information completion (Stage II). In Stage I, we first input the node and its text information into the contextual prosecutor of the LLM agent to generate context evidence $t_c^{\text{evi}}$. Subsequently, The structural prosecutor then integrates the text information of node and the structural information of the graph to generate structural evidence $t_s^{\text{evi}}$. Finally, the judge makes the final verdict $t^{\text{verd}}$ based on the context and structural evidence provided by the prosecutor and the original information of the graph. 
In phase II, the raw node text information and the final verdict output by the judge in natural language form are first converted into numerical features understandable by the graph neural network using a frozen pre-trained text encoder. Then, in each epoch, the gating mechanism provided by Eq.~\eqref{eq:gate1} and Eq.~\eqref{eq:gate2} is used to fuse the original node features $\mathbf{x}_v^\textrm{orig}$ and the verdict features $\mathbf{x}_v^\textrm{verd}$, and the fused features $\mathbf{h}_v$ are input into the GNN together with the structure of the graph $\mathbf{A}$ to update the node representation. The loss $\mathcal{L}$ is minimized according to Eq.~\eqref{eq:loss}, so as to update the parameters of the gating and GNN accordingly. Finally, based on the trained model parameters, the final node anomaly score $f_{\textrm{score}}\left (\mathcal{V} \right )$ is computed using Eq.~\eqref{eq:inf}.

\section{More Experimental Setup}
\subsection{Datasets details}\label{sec:appendix_datasets}
We evaluate our proposed framework on four widely used benchmark datasets: the citation networks Cora, Pubmed, and ogbn-Arxiv, as well as the History e-commerce network. The detailed descriptions of four datasets are as follows:

\textbf{Citation Networks}. Cora, Pubmed, and ogbn-Arxiv are citation networks~\cite{sen2008collective,hu2020open} in which nodes represent academic papers and edges indicate citation information between these papers. The node attributes encompass the titles and abstracts of research papers.

\textbf{E-commerce Networks}. History dataset~\cite{yan2023comprehensive} is extracted from the Amazon dataset~\cite{ni2019justifying}, where nodes represent various types of items, edges signify items that are frequently purchased or browsed together, and the node attributes are derived from the titles and descriptions of the respective books.

To address the absence of explicitly labeled anomalies in existing text-attributed graph datasets, we follow standard construction methods from prior studies~\cite{ding2019deep,liu2021anomaly,duan2023graph,duan2023graph}  to develop a tailored anomaly labeling system, specifically designed for text-attributed graph anomaly detection, and apply it to adjust publicly available datasets. The total number of anomalies for each dataset is presented in the final column of Table~\ref{tab:dataset_stats}.

\textbf{Contextual anomaly.} Contextual anomalies refer to nodes whose attributes are demonstrably disparate from those of their neighboring nodes~\cite{song2007conditional,ma2021comprehensive}. We design two novel strategies, insertion and replacement, to perturb the original textual attributes of nodes to generate contextual anomalies. 
To generate such anomalies, we first randomly select a target node $v_i$ and then sample a set of $K$ nodes as the candidate set. We employ the BGE~\cite{bge_embedding} to encode textual information into attribute vectors and calculate the cosine similarity between $v_i$ and each node in the candidate set. Subsequently, we select the node $v_j$ with the lowest similarity in the candidate set as the source of abnormal information. The first strategy is to insert a specified segment of text from $v_j$ into a randomly selected position within the text of $v_i$. Another approach is to randomly replace the text. This entails randomly selecting an equal number of sentences from $v_i$ and $v_j$, and replacing the corresponding sentences from $v_i$ with those from $v_j$. Both insertion and replacement strategies construct the same number of contextual anomalies. Here, we set $K = 50$ to ensure the disturbance amplitude is large enough.

\textbf{Structural anomaly.} Structural anomaly nodes usually have different connection patterns~\cite{ma2021comprehensive}, such as forming dense connections with others or connecting different communities. Therefore, we also design two strategies in this study to model these two types of structural anomalies. In real-world networks, a typical structural anomaly occurs when connections among nodes within a small clique are significantly denser than average~\cite{skillicorn2007detecting}. Thus, the first strategy injects structural anomalies that form dense connections with others. The process begins with the random selection of $q$ nodes and fully connecting them to form a clique. This step is repeated $p$ times to create $p$ such cliques, each consisting of $q$ nodes. In addition, anomaly nodes often build relationships with many benign nodes to boost their reputation and gain undue benefits, a behavior seldom seen among benign nodes~\cite{pandit2007netprobe,shin2017densealert}. Therefore, the second strategy injects structural anomalies that connect different communities by randomly adding edges. 
We start by randomly selecting a target node $v_i$. We then randomly add different numbers of edges to $v_i$ to generate structural anomalies that connect different communities. The number of edges for each target node $v_i$ is determined by sampling from the degree distribution of the original graph dataset. This approach ensures that the newly added structural anomalies continue to exhibit statistical characteristics aligned with those of the original graph. Assuming the total number of abnormal nodes is $4m$, $m$ abnormal nodes are injected for each of the aforementioned strategies.

\subsection{Implementation Details} \label{sec:appendix_param}
We report the mean and standard deviation of the results of all experiments run 5 times using different randomized seeds.
The execution environment for generating evidence and final verdicts in Stage I, including the contextual and structural prosecutors (Llama 3.1 8B) and the judge (Llama 3.1 70B)~\cite{dubey2024llama}, is as follows: Ubuntu 20.04, CPU: Intel(R) Xeon(R) Platinum 8163, GPU: NVIDIA A100 × 2, CUDA 11.6, and Memory: 500 GiB. The end-to-end training of the gating mechanism and GNN in Stage II was conducted on the following computational infrastructure: Ubuntu 22.04, CPU: AMD EPYC 7542 32-Core, GPU: NVIDIA 4090 × 1, CUDA 12.2, and Memory: 500 GiB.
In addition, versions of relevant software libraries and frameworks: Python: 3.8.13, torch: 1.12.1, torch-cluster: 1.6.0, torch-geometric: 2.1.0.post1, torch-scatter: 2.0.9, torch-sparse: 0.6.15, torch-spline-conv: 1.2.1, torchaudio: 0.12.1, torchvision: 0.13.1, transformers: 4.24.0, DGL: 0.9.0.
Finally, the range of values tried per parameter during development: the learning rate parameter is selected from \{4e-4, 5e-4, 3e-3, 4e-3, 5e-3\}, epoch is selected from \{5, 10, 25, 50, 75, 100, 200\}, weight decay is selected from \{0, 1e-3, 1e-4\}. \looseness=-1

\section{Supplementary Experimental Results}

\begin{figure}
\centering
\subfloat[sampling rounds $R$]{\label{fig:param3}\includegraphics[width=0.5\linewidth]{
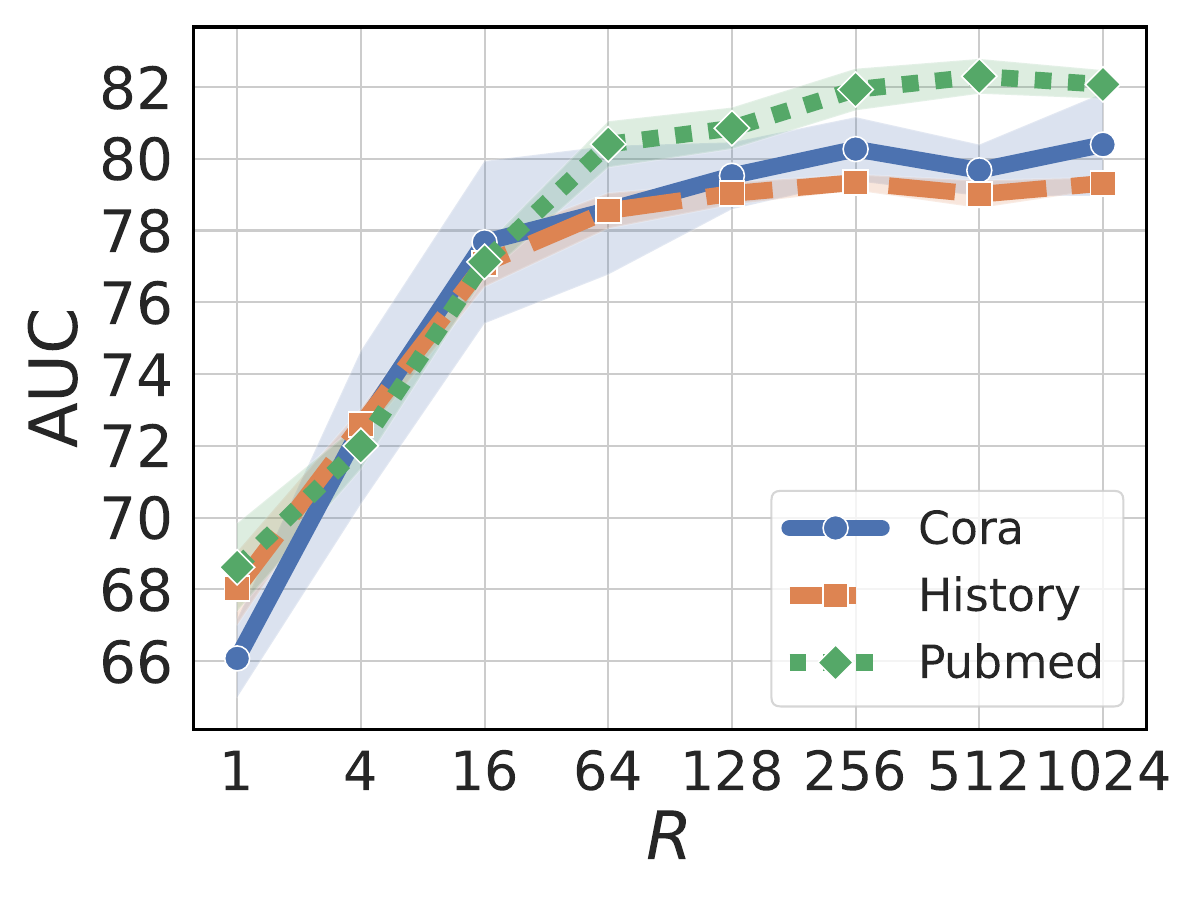}}
\subfloat[dimension $d$]{\label{fig:param4}\includegraphics[width=0.5\linewidth]{
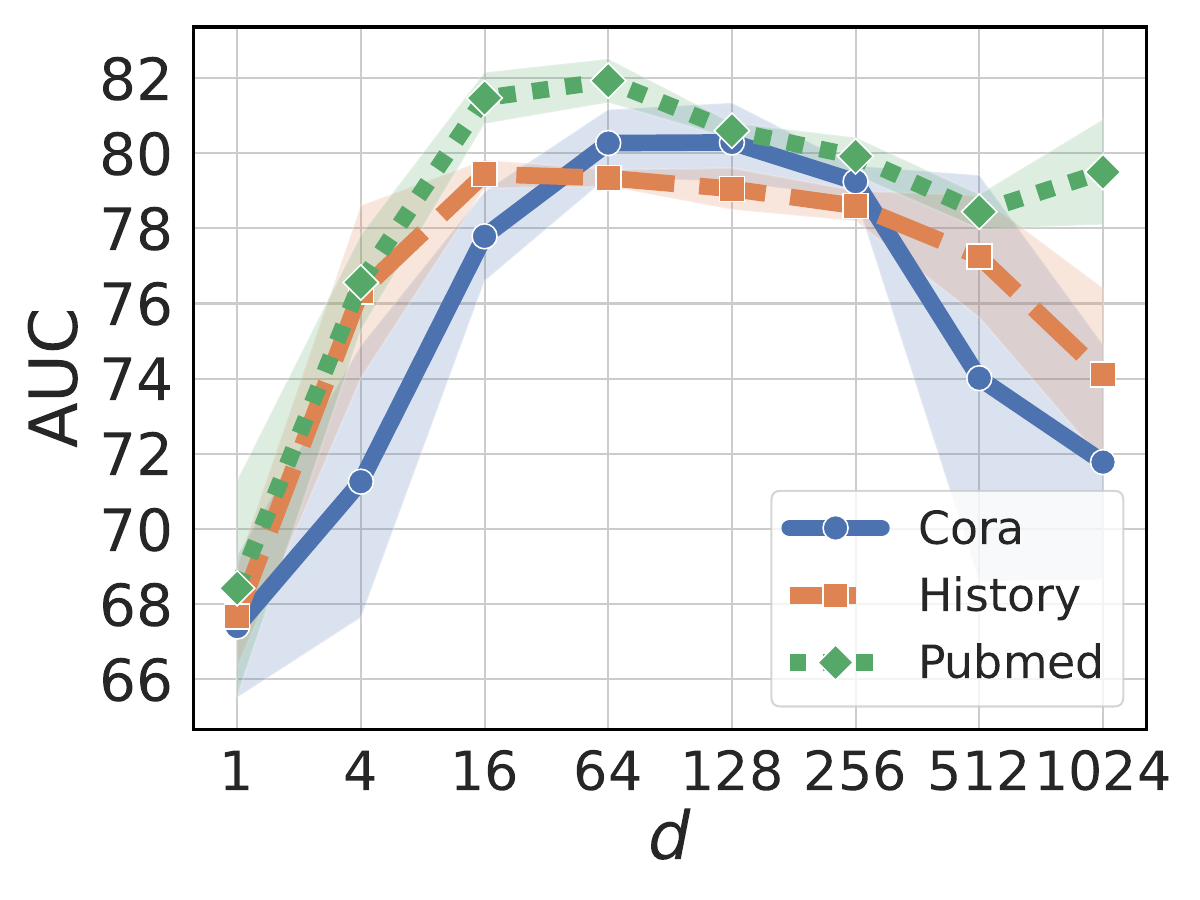}}
\caption{The parameter study of CoLL with varying (a) sampling rounds $R$, (b) dimension $d$ on the Cora, History, and Pubmed datasets w.r.t. AUC.}

\label{fig:param_appendix}
\end{figure}

\begin{figure*}
\centering
\subfloat[batch $b$ w.r.t. AUC]{\label{fig:param5}\includegraphics[width=0.25\linewidth]{
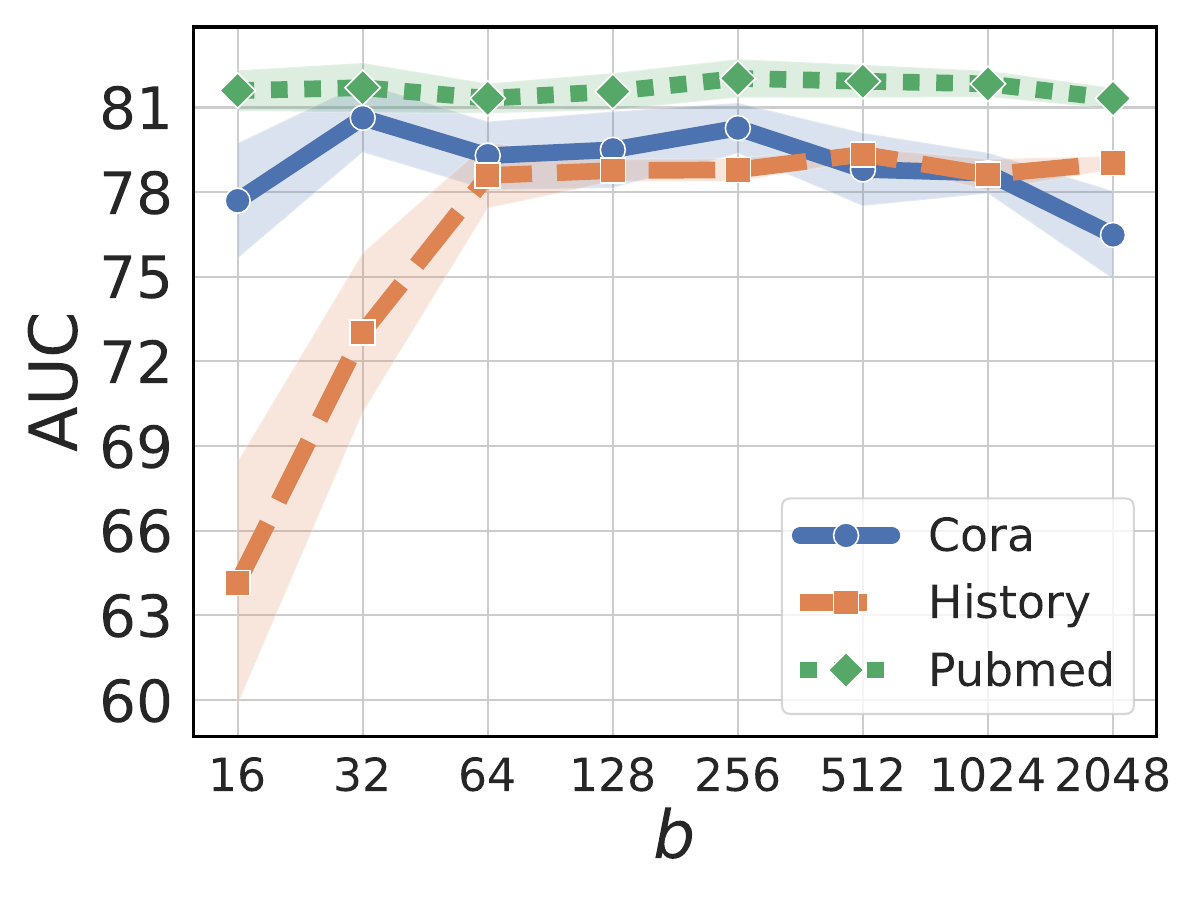}}
\subfloat[epoch $e$ w.r.t. AUC]{\label{fig:param6}\includegraphics[width=0.25\linewidth]{
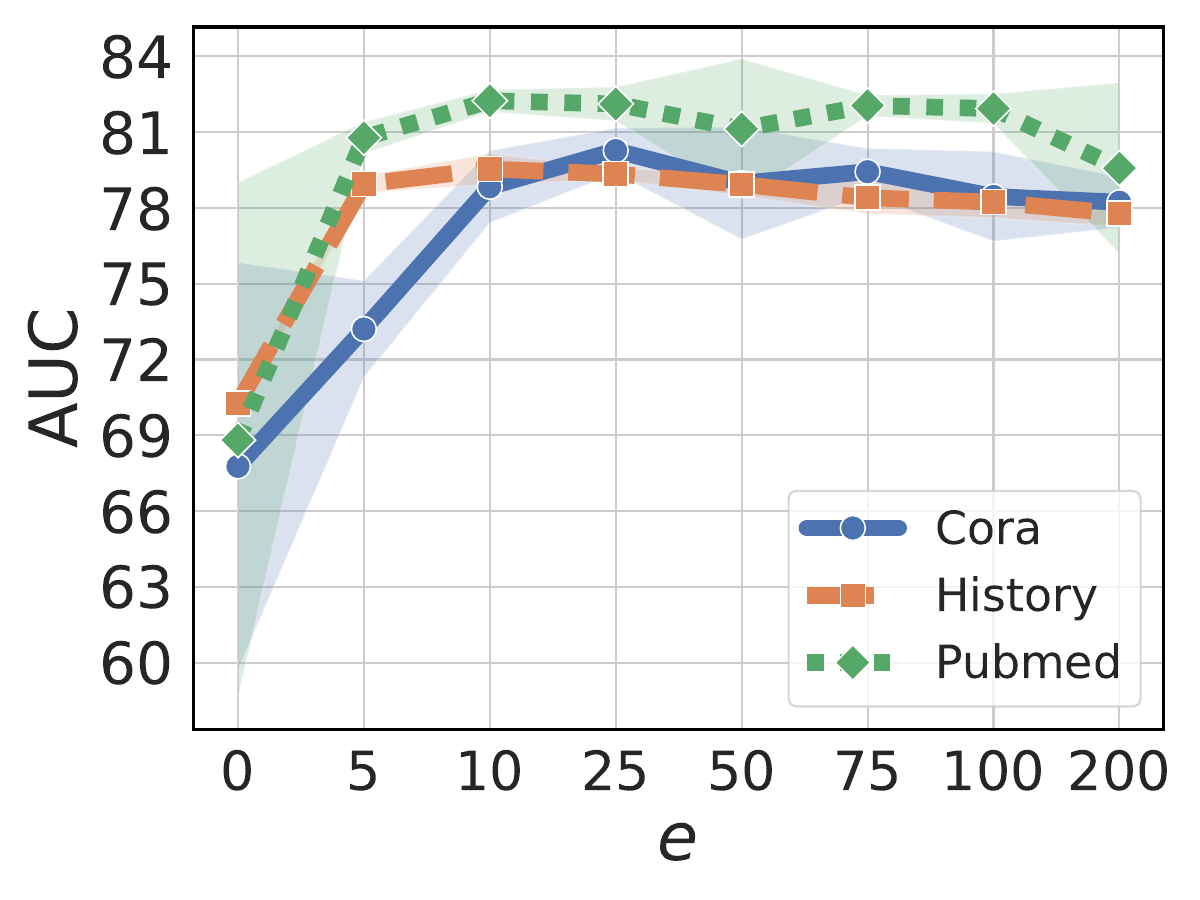}}
\subfloat[batch $b$ w.r.t. AP]{\label{fig:param7}\includegraphics[width=0.25\linewidth]{
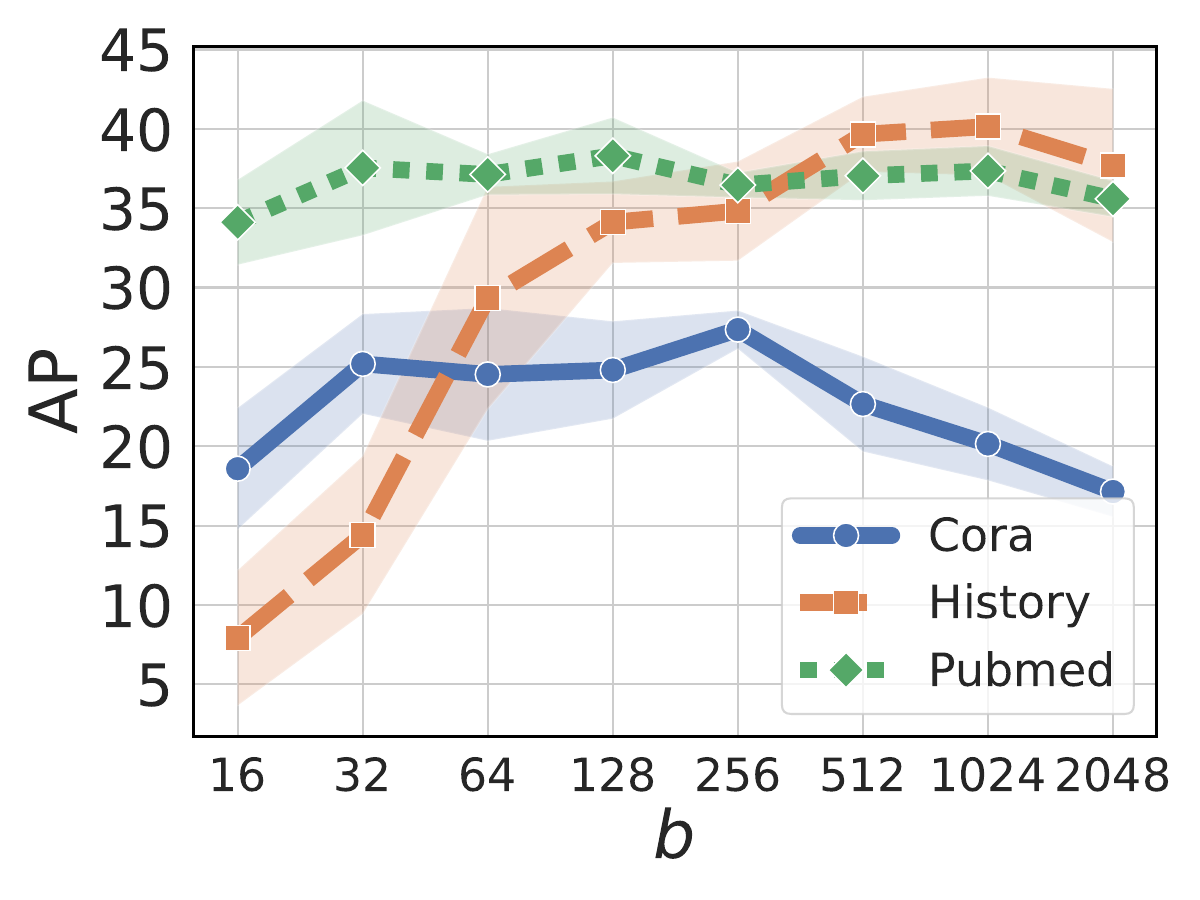}}
\subfloat[epoch $e$ w.r.t. AP]{\label{fig:param8}\includegraphics[width=0.25\linewidth]{
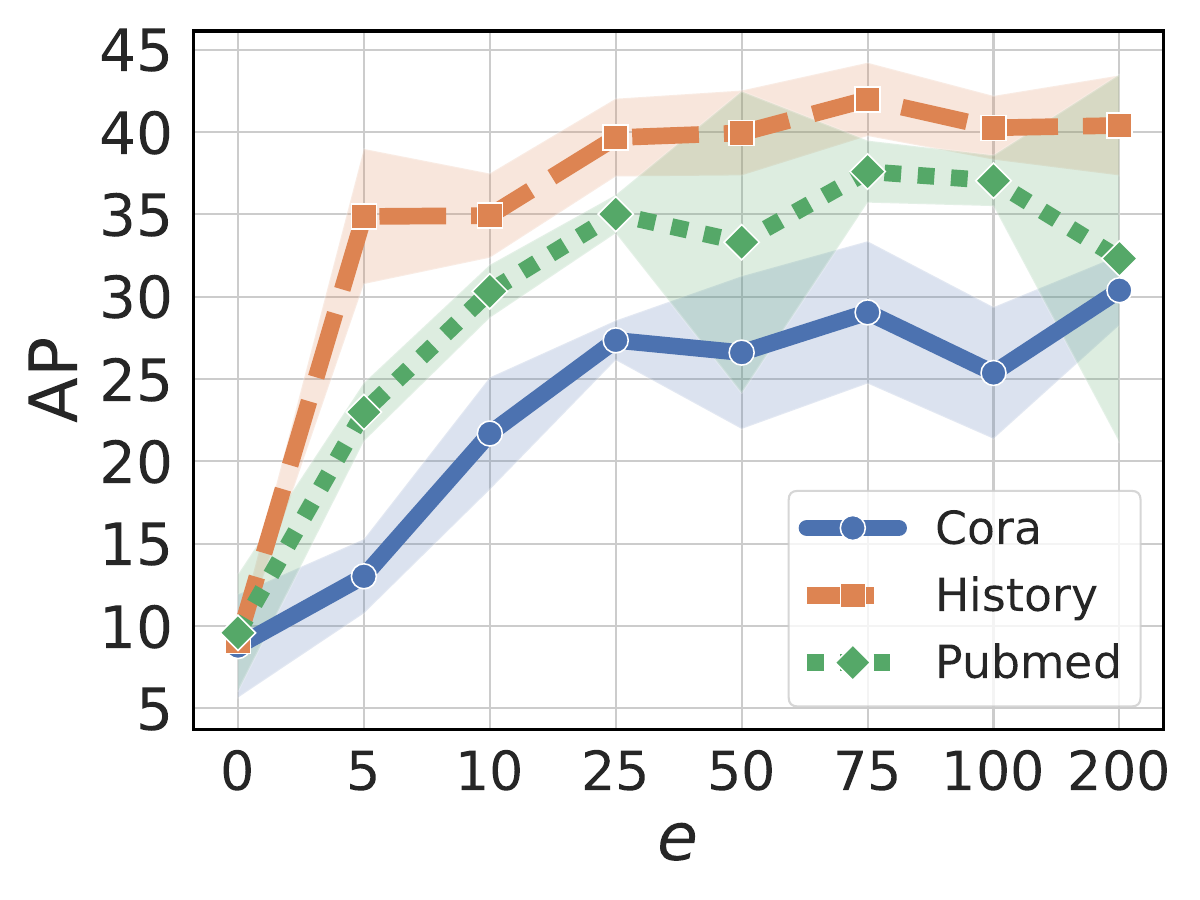}}
\caption{The parameter study of CoLL with batch $b$ and epoch $e$ on the Cora, History, and Pubmed datasets.}
\label{fig:param_appendix1}
\end{figure*}

\subsection{More Parameter Study}
\textbf{Effect of sampling rounds $R$ and hidden dimension $d$}\quad We also evaluated the effect of sampling rounds $R$ and hidden dimension $d$ on the AUC metric for CoLL. As shown in Figure~\ref{fig:param_appendix} and Figure~\ref{fig:param}, the trends for AUC and AP are consistent. For sampling rounds $R$, the improvement in AUC becomes marginal when $R$ exceeds 256. Considering the trade-off between performance and computational efficiency, we set $R$ to 256. For the hidden dimension $d$, optimal performance is typically achieved at $d=64$, as both smaller and larger dimensions result in performance degradation due to underfitting or increased noise. Therefore, we recommend setting the hidden dimension to 64 for TAGAD tasks.

\textbf{Effect of batch $b$}\quad Figure~\ref{fig:param5} and Figure~\ref{fig:param7}  illustrate the impact of batch size $b$ on AUC and AP, respectively, across the Cora, History, and Pubmed datasets. Overall, we observe that a moderate batch size is beneficial for model performance, while excessively large or small batches can degrade results. Initially, as $b$ increases from 16 to 128, both AUC and AP improve significantly across all datasets, indicating that a sufficiently large batch allows for more stable gradient updates and better generalization. Beyond this range, the performance plateaus, suggesting diminishing returns from increasing batch size. This trend is particularly noticeable in History and Pubmed, where performance stabilizes after $b=128$. A slight performance drop is observed for substantial batch sizes ($b>512$), especially in Cora, where AUC and AP decline. This is likely due to reduced gradient variance, which limits the model's ability to escape sharp minima, leading to suboptimal convergence. In conclusion, batch sizes between 64 and 512 provide an optimal trade-off between performance and efficiency, ensuring stable training dynamics and robust anomaly detection.

\textbf{Effect of epoch $e$}\quad Figure~\ref{fig:param6} and Figure~\ref{fig:param8} present the parameter sensitivity analysis of epoch $e$ with respect to AUC and AP, respectively, across the Cora, History, and Pubmed datasets. The results indicate that increasing the number of training epochs initially enhances model performance, but excessive training leads to overfitting. For small epoch values $e<25$ both AUC and AP are suboptimal across all datasets, suggesting that the model lacks sufficient training to fully learn meaningful representations. As $e$ increases from 5 to 50, performance improves significantly, with all datasets reaching near-optimal AUC and AP values. This trend highlights the importance of sufficient training epochs for convergence. Beyond $e=75$, the performance gains saturate, and a slight decline is observed, particularly on Cora and Pubmed. This indicates possible overfitting, where the model begins to memorize patterns in the training data rather than generalizing to anomalies effectively. Overall, setting $e$ between 25 and 100 provides a good trade-off between stability and generalization, ensuring robust anomaly detection without overfitting.

\begin{table}[t]
\renewcommand\arraystretch{1.1}
\centering
\caption{Running time comparison (in seconds) on Pubmed and History, including average time and ranking.}
\label{tab:runtime}
\begin{tabular}{lcccc}
    \toprule
    \textbf{Method} & \textbf{Pubmed} & \textbf{History} & \textbf{Avg. Time (s)} & \textbf{Rank} \\
    \midrule
    \multicolumn{5}{c}{\textit{Traditional Methods}} \\
    \midrule
    LOF & 1.44 & 4.48 & 2.96 & 2 \\
    SCAN & 107.52 & 561.90 & 334.71 & 5 \\
    Radar & 22.96 & 195.07 & 109.02 & 4 \\
    \midrule
    \multicolumn{5}{c}{\textit{Deep Learning-Based Methods}} \\
    \midrule
    AEGIS & 1,552.91 & 3,232.93 & 2,392.92 & 10 \\
    MLPAE & 49.48 & 65.57 & 57.53 & 3 \\
    DOMINANT & 0.46 & 0.65 & 0.56 & \textbf{1} \\
    GAD-NR & 14,008.06 & 35,196.88 & 24,602.47 & 12 \\
    \midrule
    CoLA & 1,052.65 & 2,587.15 & 1,819.90 & 7 \\
    ANEMONE & 1,101.61 & 2,704.56 & 1,903.08 & 8 \\
    SL-GAD & 2,231.07 & 5,317.08 & 3,774.08 & 11 \\
    GRADATE & 1,408.88 & 3,375.16 & 2,392.02 & 9 \\
    \midrule
    \rowcolor[HTML]{E9E9E9}
    CoLL & 467.81 & 517.40 & 492.61 & 6 \\
    \bottomrule
\end{tabular}
\end{table}

\subsection{Time Complexity}
Besides Figure~\ref{fig:time}, we present the detailed running time (in seconds) of our method CoLL and various baselines on the Pubmed and History datasets. As shown in Table~\ref{tab:runtime}, traditional methods typically exhibit high computational efficiency. However, Table~\ref{tab:ad} shows they often lack strong anomaly detection capabilities. Deep learning-based methods generally incur higher computational costs. Although DOMINANT achieves exceptional speed, its requirement to reconstruct the entire graph makes it difficult to scale to large graphs. \looseness=-1

Our method ranks 3rd among the 9 deep learning-based approaches in terms of computational efficiency and runs 9.37× faster on average than other deep learning methods. Meanwhile, our method achieves a 13.37\% average improvement in AP over the second-best method. This demonstrates that CoLL effectively balances accuracy and efficiency, making it a highly practical solution for scalable and effective anomaly detection.

\subsection{Case Study} \label{sec:appendix_case}
Prior GAD methods are predominantly black-box models~\cite{ma2021comprehensive}, whether based on reconstruction or contrastive learning paradigms. These methods train GAD models under the guidance of anomaly detection objectives, ultimately producing scalar anomaly scores during inference. However, the high-dimensional embeddings learned in training are inherently uninterpretable to humans, and the final anomaly scores provide little insight into the reasoning behind detections. In contrast, LLM-generated evidence is presented as human-readable rationales, further enhancing the interpretability of traditional black-box GAD approaches.

To intuitively demonstrate the effectiveness and interpretability of LLM collaboration in generating high-quality anomaly-specific evidence and the verdict, we present three representative cases across different datasets: detecting contextual anomalies, detecting structural anomalies, and handling prosecutor failure scenarios.

\begin{figure*}
  \centering
  \setlength{\belowcaptionskip}{-0.15cm}
  \includegraphics[width=0.9\linewidth]{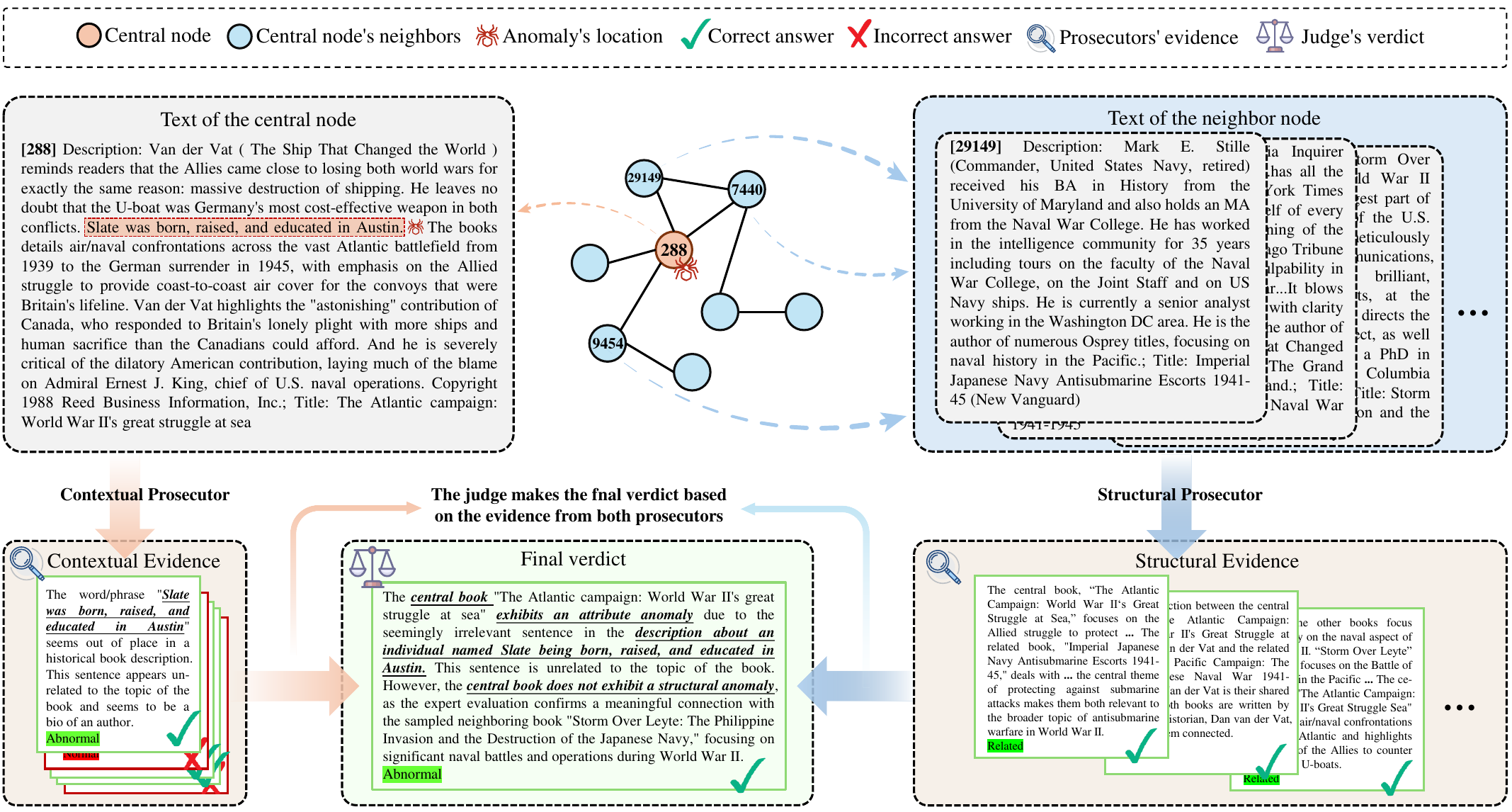}
  \caption{Case study of multi-LLM collaboration for detecting contextual anomalies in the History dataset. Node 288 presents a contextual anomaly without structural anomaly. The contextual prosecutor, relying solely on raw text, detects most contextual anomalies but lacks full accuracy. The structural prosecutor finds no edge anomalies. The judge refines the contextual prosecutor’s evidence, accurately identifying the anomaly and highlighting its specific anomalous parts with explanations.}
  \label{fig:case-attr}
\end{figure*}

\textbf{Scenarios of detecting contextual anomalies}\quad As shown in Figure~\ref{fig:case-attr}, Node 288 from the History dataset is connected to multiple edges, including (288, 29149), (288, 7440), and (288, 9454), etc. Node 288 exhibits an contextual anomaly, highlighted in the red box within the node's text, while no structural anomalies are observed on its edges. The contextual prosecutor, based solely on the raw text of the node, identifies contextual anomalies in 3 out of 5 outputs, with 2 outputs indicating no anomaly. The structure prosecutor, relying on the raw text of the node and its neighbors, detects no edge anomalies. By integrating the multiple pieces of evidence from both prosecutors, the judge conducts further analysis to deliver a detailed verdict, successfully identifying the contextual anomaly in Node 288, along with its specific anomalous parts and explanations.

\begin{figure*}
  \centering
  \setlength{\belowcaptionskip}{-0.15cm}
  \includegraphics[width=0.9\linewidth]{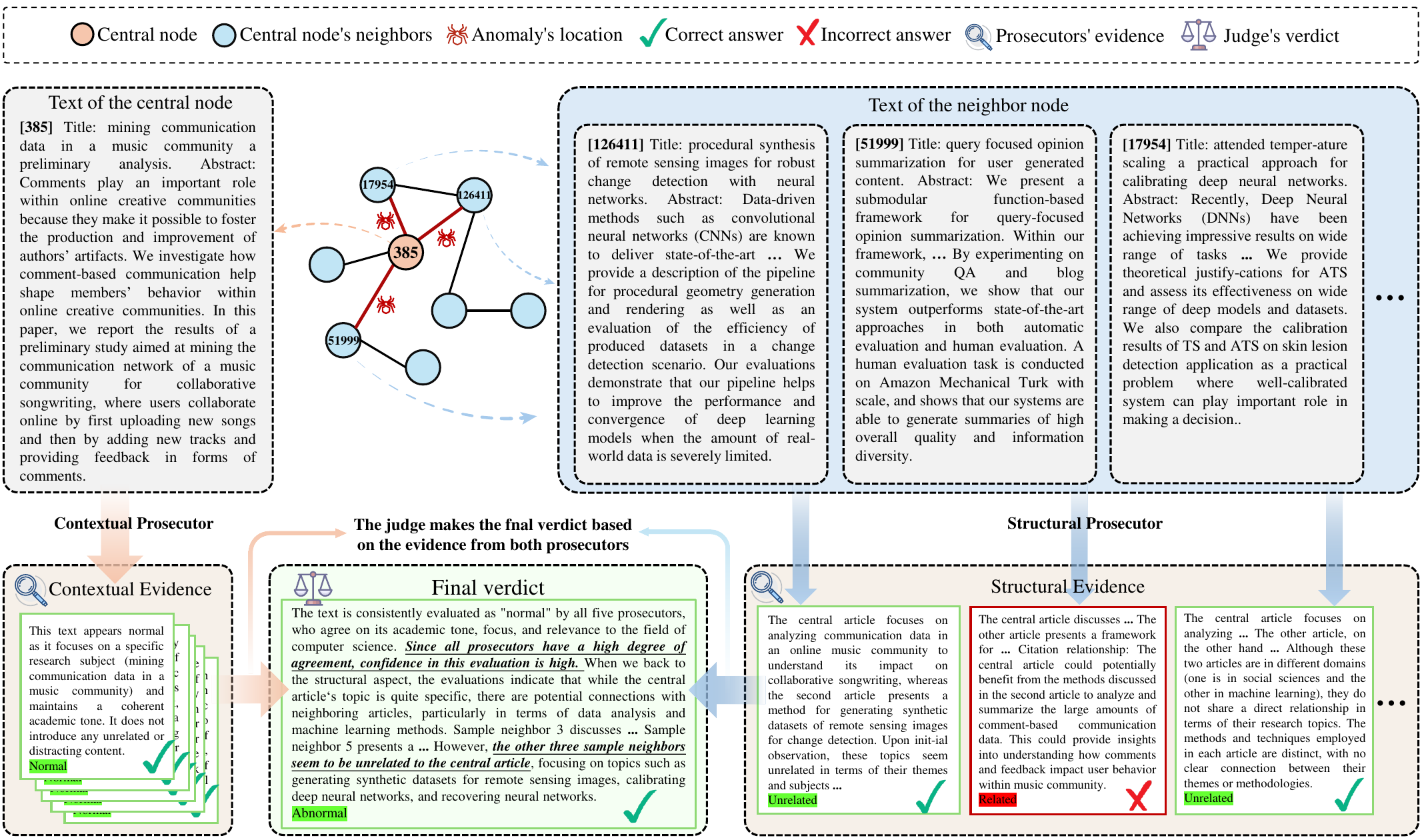}
  \caption{Case study of multi-LLM collaboration for detecting structural anomalies in the ogbn-Arxiv dataset. Node 385 has no contextual anomalies, but edges such as (385, 126411), (385, 17954), and (385, 51999) are anomalous. The contextual prosecutor deems attributes normal, while the structural prosecutor detects most edge anomalies but is not always accurate. The judge refines the structural prosecutor’s evidence, identifying the structural anomalies and pinpointing the three anomalous edges.}
  \label{fig:case-stru}
\end{figure*}

\textbf{Scenarios of detecting structural anomalies}\quad
As shown in Figure~\ref{fig:case-stru}, Node 385 from the ogbn-Arxiv dataset exhibits normal attributes but anomalous edges, such as (385, 126411), (385, 17954), and (385, 51999). The contextual prosecutor, based solely on the raw text of the node, correctly identifies no contextual anomalies in all 5 outputs. The structure prosecutor detects anomalies in 2 out of the 3 anomalous edges but misclassifies one as normal. By analyzing the raw text of all nodes and integrating evidence from both prosecutors, the judge accurately identifies Node 385 as anomalous and successfully pinpoints all three anomalous edges, effectively addressing occasional inaccuracies in the prosecutor's assessment.

\begin{figure*}
  \centering
  \includegraphics[width=0.9\linewidth]{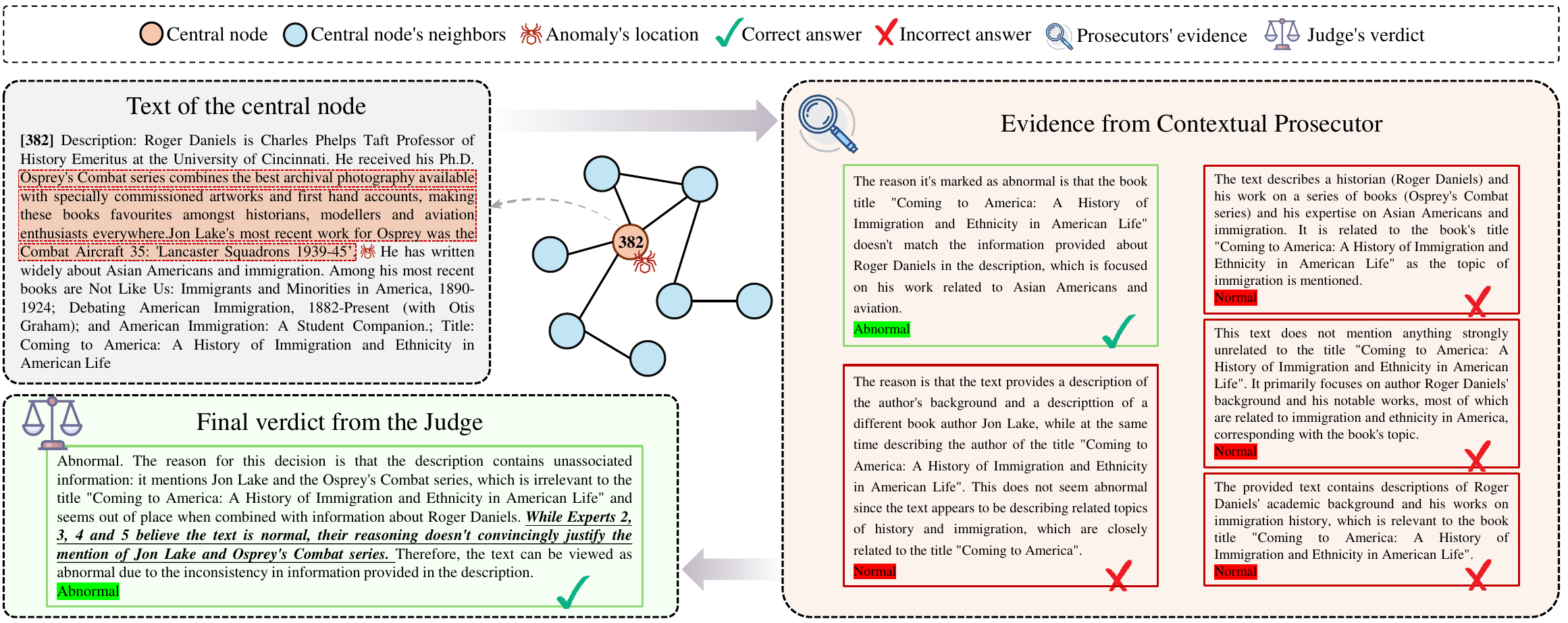}
  \caption{A representative case study in the History dataset where multi-LLM collaboration successfully detects anomalies despite prosecutor failures. Node 382 exhibits an contextual anomaly. However, in 4 out of 5 outputs, the contextual prosecutor incorrectly considers the text normal. By reviewing the node's raw text and analyzing the evidence provided by the prosecutors, the judge finds the reasoning of the other four prosecutors unconvincing and ultimately makes an accurate decision, successfully identifying the contextual anomaly in Node 382 along with an explanation. \looseness=-1}
  \label{fig:case-2phase}
\end{figure*}

\textbf{Scenarios of handling prosecutor failures}\quad
As shown in Figure~\ref{fig:case-2phase}, Node 382 from the History dataset exhibits an contextual anomaly, highlighted in the red box within the node's text. In this case, we focus exclusively on the contextual prosecutor's performance and the challenges of handling false negatives. Among the 5 pieces of evidence produced by the contextual prosecutor, only 1 correctly identifies the anomaly, while the other 4 incorrectly report no anomalies in the text. This scenario demonstrates the critical role of the judge, as heuristic rules alone would struggle to handle such cases effectively.

Despite the overwhelming majority of the prosecutors reporting no anomaly in the text, the judge carefully reviews the raw text of Node 382 and all the evidence provided. Ultimately, the judge deems the reasoning of the single anomalous report compelling and adopts it as the basis for a correct decision. This ensures that the contextual anomaly in Node 382 is successfully identified, along with a clear explanation, underscoring the indispensable value of the judge's role in such situations.

\begin{figure*}
  \centering
  \includegraphics[width=0.9\linewidth]{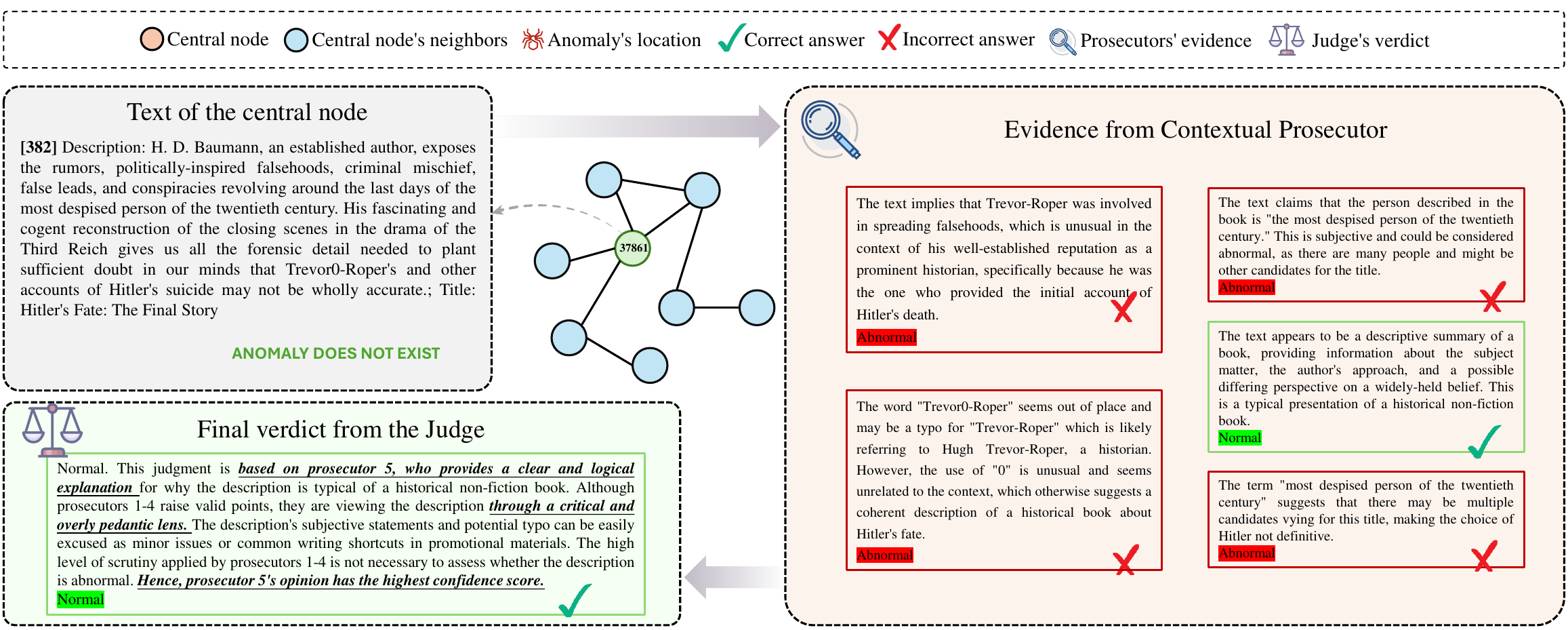}
  \caption{A representative case study in the History dataset where multi-LLM collaboration correctly identifies a normal node despite multiple prosecutors marking it as abnormal. Node 37861 does not exhibit an anomaly, but several contextual prosecutors incorrectly flag it as abnormal due to perceived issues such as a potential typo ("Trevor0-Roper") and subjective phrasing. While the prosecutors provide valuable detailed analyses, the judge's role in carefully weighing these details is equally crucial. This case highlights the importance of both thorough prosecutors analysis and a higher-level judgment that considers the overall context in anomaly detection. \looseness=-1}
  \label{fig:case-3phase}
\end{figure*}

Similarly, in Figure~\ref{fig:case-3phase}, Node 37861 presents the opposite challenge—most prosecutors incorrectly classify it as abnormal due to perceived textual issues such as a potential typo and subjective phrasing, despite the text being a typical historical book description. This case highlights the challenge of handling prosecutor disagreements. While several prosecutors flag the text as anomalous, others correctly recognize its coherence within the given context. In this scenario, the judge plays an equally crucial role—not by selecting a minority dissenting voice, as in Figure 10, but by carefully weighing all arguments and determining that the most logical and well-supported reasoning is provided by prosecutor 5.

By considering both prosecutor failures and disagreements, these cases illustrate the limitations of heuristic rules and the necessity of a judge who can critically evaluate evidence beyond simple majority voting. Whether ensuring an overlooked anomaly is detected or preventing an unjustified anomaly classification, the judge’s decision-making process is essential for accurate anomaly identification. \looseness=-1

\begin{table*}[h]
    \centering
    \renewcommand{\arraystretch}{1.5} 
    \caption{Prompts for ArXiv Anomaly Detection}
    \label{tab:prompt-arxiv}
    \begin{tabular}{p{4cm} p{13cm}}
        \toprule
        \textbf{Prompt Type} & \textbf{Example Prompt} \\
        \midrule
        \textbf{Contextual Prosecutor} & <SYS> You will get an excerpt from an academic paper submitted to ArXiv, typically within the field of computer science. Normal texts should focus on relevant research topics such as machine learning, artificial intelligence, and related areas. You need to read the excerpt carefully to see if any parts of the text introduce content that is irrelevant to the current research subject or significantly deviate from the main academic focus.\newline <USER> Is there anything unusual in the following text? Here is the text: <text attribute>\newline <SYS> You must reply with only one word "normal" or "abnormal" then provide a concise explanation of the decision. Note that you can report abnormal only if you find that some part of the review appears strongly inconsistent or irrelevant to the overall meaning of the text. Ignore spelling, grammar as long as the sentences relate to each other and maintain a common focus.\\
        \textbf{Structural Prosecutor} & <SYS> Here are the titles and abstracts of two articles from the Arxiv dataset. Your task is to determine whether there should be a citation relationship between the two articles. Focus on whether their themes, topics, or described subjects are related or have any other valid reason for citation.\newline <USER> Here is the central text: <text attribute> And this is another text <neighbor text> \newline <SYS> Provide concise explanation for the citation between the central article and the other one. Then, on a new line, if a citation relationship exists, you must strictly conclude with "related", otherwise, report "unrelated". \\
        \textbf{Judge Phase} & <SYS> You are provided with a central academic paper from the ArXiv dataset, along with: Evaluations from five prosecutors assessing whether the central paper's content or abstract contains contextual anomalies, such as irrelevant content or significant deviation from its stated research domain. Evaluations from five prosecutors assessing whether the central paper has meaningful topical connections with its sampled neighboring papers (structural anomalies). Note that in the structural evaluations, some sampled pairs may overlap, allowing for comparisons of judgment consistency. These prosecutor opinions may vary in reliability. Your task is to carefully analyze their consistency and correctness to judge their overall credibility. Based on this analysis, and by independently reviewing the central paper’s abstract and its sampled connections where necessary, determine if the central paper exhibits any anomalies: Contextual anomaly: Assess if the central paper’s abstract significantly deviates from its stated research field or includes irrelevant information. Structural anomaly: Assess if the central paper has at least one incorrectly related or unrelated neighboring paper, leveraging repeated samples to evaluate consistency where applicable. If the central paper has no neighbors, conclude that there is no structural anomaly. If any anomaly (attribute or structural) is identified, conclude that the central paper is "Abnormal." If no anomalies are detected, conclude that it is "Normal." \newline <USER> Here is the central text: <text attribute> and the following are opinions from 10 prosecutors: for <corresponding text> prosecutor: <prosecutor opinion>. \newline <SYS> Provide a BRIEF summary explaining whether the central book exhibits any anomalies, integrating all the information provided and considering the consistency and correctness of the evaluations. On a NEW line, choose ONLY ONE WORD either "Normal" or "Abnormal" as your final judgment. \\
        \bottomrule
    \end{tabular}
\end{table*}

\begin{table*}[h]
    \centering
    \renewcommand{\arraystretch}{1.5} 
    \caption{Prompts for History Anomaly Detection}
    \label{tab:prompt-history}
    \begin{tabular}{p{4cm} p{13cm}}
        \toprule
        \textbf{Prompt Type} & \textbf{Example Prompt} \\
        \midrule
        \textbf{Contextual Prosecutor} & <SYS> You will get a piece of text about a historical book, which contains the Description of the book and its Title. You need to read the text sentence by sentence and determine whether the text is abnormal. \newline <USER> Is there anything unusual in the following text? Here is the text: <text attribute>\newline <SYS> You must reply with only one word "normal" or "abnormal" then provide a concise explanation of the decision. Note that you can report abnormal only if you find that some part of the review appears strongly inconsistent or irrelevant to the overall meaning of the text. Ignore spelling, grammar as long as the sentences relate to each other and maintain a common focus.\\
        \textbf{Structural Prosecutor} & <SYS> You are given two books, each with a title and description. Normally, two history books are connected only if their content is clearly related. Your task is to determine whether these two books should be considered related based on their content. Ignore minor issues like grammar. Provide a brief, clear explanation of why this connection is appropriate or not. \newline <USER> Here is the central book: <text attribute> And this is another book <neighbor text> \newline <SYS> Provide brief explanation for the connection between the central book and the related books. Then, on a new line, give your final judgment: "related" or "unrelated." \\
        \textbf{Judge Phase} & <SYS> You are provided with a central book from the History dataset, along with: Evaluations from five prosecutors assessing whether the central book's content or description contains contextual anomalies, such as irrelevant content or significant deviation from its historical subject. Evaluations from five prosecutors assessing whether the central book has meaningful thematic connections with its sampled neighboring books (structural anomalies). Note that in the structural evaluations, some sampled pairs may overlap, allowing for comparisons of judgment consistency. These prosecutor opinions may vary in reliability. Your task is to carefully analyze their consistency and correctness to judge their overall credibility. Based on this analysis, and by independently reviewing the central book's description and its sampled connections where necessary, determine if the central book exhibits any anomalies: Contextual anomaly: Assess if the central book's description significantly deviates from its historical subject or includes irrelevant information. Structural anomaly: Assess if the central book has at least one incorrectly related or unrelated neighboring book, leveraging repeated samples to evaluate consistency where applicable. If the central book has no neighbors, conclude that there is no structural anomaly. If any anomaly (attribute or structural) is identified, conclude that the central book is "Abnormal." If no anomalies are detected, conclude that it is "Normal." \newline <USER> Here is the central book: <text attribute> and the following are opinions from 10 prosecutors: for <corresponding text> prosecutor: <prosecutor opinion>. \newline <SYS> Provide a BRIEF summary explaining whether the central book exhibits any anomalies, integrating all the information provided and considering the consistency and correctness of the evaluations. On a NEW line, choose ONLY ONE WORD either "Normal" or "Abnormal" as your final judgment. \\
        \bottomrule
    \end{tabular}
\end{table*}

\begin{table*}[h]
    \centering
    \renewcommand{\arraystretch}{1.5} 
    \caption{Prompts for Cora Anomaly Detection}
    \label{tab:prompt-cora}
    \begin{tabular}{p{4cm} p{13cm}}
        \toprule
        \textbf{Prompt Type} & \textbf{Example Prompt} \\
        \midrule
        \textbf{Contextual Prosecutor} & <SYS> You will get an excerpt from an academic paper in the field of computer science. Normal texts should focus on the current research topic, typically related to fields like machine learning, artificial intelligence, or related subdomains. You need to read the excerpt carefully to see if any parts of the text introduce content that is irrelevant to the current computer science subject or deviate significantly from the main research theme. \newline <USER> Is there anything unusual in the following text? Here is the text: <text attribute>\newline <SYS> You must reply with only one word "normal" or "abnormal" then provide a concise explanation of the decision. Note that you can report abnormal only if you find that some part of the review appears strongly inconsistent or irrelevant to the overall meaning of the text. Ignore spelling, grammar as long as the sentences relate to each other and maintain a common focus. Let's think step by step.\\
        \textbf{Structural Prosecutor} & <SYS> Here are the titles and, if available, the abstracts from two articles in the CORA dataset. Please analyze the texts carefully to determine whether there is a citation relationship between the articles. Provide a brief, clear explanation of why the connection is appropriate or not. \newline <USER> Here is the central text: <text attribute> And this is another text <neighbor text> \newline <SYS> Provide concise explanation for the citation between the central article and the other one. Then, on a new line, if a citation relationship exists, you must strictly conclude with "related", otherwise, report "unrelated". \\
        \textbf{Judge Phase} & <SYS> You are provided with a central academic paper from the Cora dataset, along with: Evaluations from five prosecutors assessing whether the central paper's content or abstract contains contextual anomalies, such as irrelevant content or significant deviation from its stated research domain. Evaluations from five prosecutors assessing whether the central paper has meaningful topical connections with its sampled neighboring papers (structural anomalies). Note that in the structural evaluations, some sampled pairs may overlap, allowing for comparisons of judgment consistency. These prosecutor opinions may vary in reliability. Your task is to carefully analyze their consistency and correctness to judge their overall credibility. Based on this analysis, and by independently reviewing the central paper’s abstract and its sampled connections where necessary, determine if the central paper exhibits any anomalies: Contextual anomaly: Assess if the central paper’s abstract significantly deviates from its stated research field or includes irrelevant information. Structural anomaly: Assess if the central paper has at least one incorrectly related or unrelated neighboring paper, leveraging repeated samples to evaluate consistency where applicable. If the central paper has no neighbors, conclude that there is no structural anomaly. If any anomaly (attribute or structural) is identified, conclude that the central paper is "Abnormal." If no anomalies are detected, conclude that it is "Normal." \newline <USER> Here is the central text: <text attribute> and the following are opinions from 10 prosecutors: for <corresponding text> prosecutor: <prosecutor opinion>. \newline <SYS> Provide a BRIEF summary explaining whether the central book exhibits any anomalies, integrating all the information provided and considering the consistency and correctness of the evaluations. On a NEW line, choose ONLY ONE WORD either "Normal" or "Abnormal" as your final judgment. \\
        \bottomrule
    \end{tabular}
\end{table*}

\begin{table*}[h]
    \centering
    \renewcommand{\arraystretch}{1.5} 
    \caption{Prompts for Pubmed Anomaly Detection}
    \label{tab:prompt-pubmed}
    \begin{tabular}{p{4cm} p{13cm}}
        \toprule
        \textbf{Prompt Type} & \textbf{Example Prompt} \\
        \midrule
        \textbf{Contextual Prosecutor} & <SYS> You will get an excerpt from an academic paper in the medical or biomedical field. Normal texts should focus on the current medical research topic. You need to read the excerpt carefully to see if any parts of the text introduce content that is irrelevant to the current medical subject or deviate significantly from the main research theme.\newline <USER> Is there anything unusual in the following text? Here is the text: <text attribute>\newline <SYS> You must reply with only one word "normal" or "abnormal" then provide a concise explanation of the decision. Note that you can report abnormal only if you find that some part of the review appears strongly inconsistent or irrelevant to the overall meaning of the text. Ignore spelling, grammar as long as the sentences relate to each other and maintain a common focus.\\
        \textbf{Structural Prosecutor} & <SYS> Here are the titles and, if available, the abstracts from two articles in the PubMed dataset. Please analyze the texts carefully to determine whether there is a citation relationship between the articles. Provide a brief, clear explanation of why the connection is appropriate or not. \newline <USER> Here is the central text: <text attribute> And this is another text <neighbor text> \newline <SYS> Provide concise explanation for the citation between the central article and the other one. Then, on a new line, if a citation relationship exists, you must strictly conclude with "related", otherwise, report "unrelated". \\
        \textbf{Judge Phase} & <SYS> You are provided with a central article from the PubMed dataset, along with: Results from five evaluations assessing whether the text of the central article contains contextual anomalies, such as irrelevant content or deviation from its main research theme. Results from five evaluations assessing whether the central article has meaningful citation relationships with its sampled neighboring articles (structural anomalies). Note that in the structural evaluations, some sampled pairs may overlap, allowing for comparisons of judgment consistency. Your task is to determine if the central article exhibits any anomalies: Contextual anomaly: Assess if the central article’s content significantly deviates from its main theme or includes irrelevant information. Structural anomaly: Assess if the central article has at least one incorrectly related or unrelated neighboring article, leveraging repeated samples to evaluate consistency where applicable. If any anomaly (attribute or structural) is identified, conclude that the central article is "Abnormal". If no anomalies are detected, conclude that it is "Normal". On a new line, provide a single-word judgment: "Normal" if no anomaly is detected. "Abnormal" if any anomaly is identified. \newline <USER> Here is the central text: <text attribute> and the following are opinions from 10 prosecutors: for <corresponding text> prosecutor: <prosecutor opinion>. \newline <SYS> Provide a concise and comprehensive summary explaining whether the central article exhibits any anomalies, integrating all the information provided. Clearly specify if the anomaly is related to the article's content (contextual anomaly) or its relationships with neighboring articles (structural anomaly). On a new line, give your final judgment: "Normal" if no anomaly is detected. "Abnormal" if any anomaly is identified. \\
        \bottomrule
    \end{tabular}
\end{table*}

\section{Prompt Design} \label{sec:appendix_prompt}
Tables~\ref{tab:prompt-arxiv}, ~\ref{tab:prompt-history}, ~\ref{tab:prompt-cora} and ~\ref{tab:prompt-pubmed} outline the prompts used for anomaly detection in the four datasets. Each prompt is designed to capture specific types of anomalies by leveraging structured, domain-specific queries that guide the language model in decision-making. The contextual prosecutor prompt evaluates content relevance, the structural prosecutor prompt examines citation consistency, and the judge integrates multiple prosecutor assessments to determine the overall anomaly classification. Each prompt guides the model by explicitly instructing it to analyze specific elements and provide a clear, structured response. The model generates an answer in a predefined format, ensuring interpretability and ease of extraction. 

\section{Limitations}
In this study, we exclusively employed Llama~\cite{dubey2024llama} for multi-LLM collaboration. While evaluating a broader range of state-of-the-art LLMs might provide additional insights, our focus is not on LLM selection but rather on leveraging LLMs to extract anomaly-specific evidence. This work lays the foundation for more advanced LLM-driven designs for text-attributed graph anomaly detection.
More powerful LLMs in the future can be seamlessly integrated into our framework to further enhance performance. Additionally, considering the sensitive nature of GAD applications, such as fraud detection and cybersecurity, we intentionally avoided API-based online language generation interfaces, which might offer better performance but pose a risk of data leakage. Finally, while CoLL provides a strong framework, it is essential to avoid over-reliance on fully automated anomaly detection systems for high-stakes decisions. A hybrid approach combining automated detection with human oversight is advisable to mitigate potential risks.

\end{document}